\documentclass[acmsmall, authorversion, nonacm]{acmart}

\usepackage{algorithm}
\usepackage{multirow}
\usepackage[noend]{algpseudocode}
\usepackage{comment}
\usepackage{subfigure}

\setcopyright{acmcopyright}
\acmJournal{TODAES}
\acmYear{2019}\acmVolume{00}\acmNumber{0}\acmArticle{000}\acmMonth{9}
\acmDOI{10.1145/1122445.1122456}

\usepackage{framed}

\usepackage{soul}

\usepackage{tikz}
\newcommand*\circled[1]{\tikz[baseline=(char.base)]{
            \node[shape=circle,draw,inner sep=2pt] (char) {#1};}}

\begin{document}

%
% The "title" command has an optional parameter, allowing the author to define a "short title" to be used in page headers.
\title[Are Adversarial Perturbations a Showstopper for ML-Based CAD?]{Are Adversarial Perturbations a Showstopper for ML-Based CAD? A Case Study on CNN-Based Lithographic Hotspot Detection}
% \titlenote{\textbf{Submitted to the Special Issue on Machine Learning for CAD (ML-CAD)}}

%
% The "author" command and its associated commands are used to define the authors and their affiliations.
% Of note is the shared affiliation of the first two authors, and the "authornote" and "authornotemark" commands
% used to denote shared contribution to the research.
\author{Kang Liu}
\email{kang.liu@nyu.edu}
\affiliation{%
  \institution{New York University}
  \department{Center for Cybersecurity}
  \streetaddress{2 Metrotech Center}
  \city{New York}
  \state{NY}
  \postcode{11201}
  \country{USA}
}

\author{Haoyu Yang}
\email{hyyang@cse.cuhk.edu.hk}
\author{Yuzhe Ma}
\email{yzma@cse.cuhk.edu.hk}
\affiliation{%
  \institution{Chinese University of Hong Kong}
%   \department{}
%   \streetaddress{}
  \city{Hong Kong}
%   \state{}
%   \postcode{}
%   \country{}
}

\author{Benjamin Tan}
\email{benjamin.tan@nyu.edu}
\orcid{0002-7642-3638}
\affiliation{%
  \institution{New York University}
  \department{Center for Cybersecurity}
  \streetaddress{2 Metrotech Center}
  \city{New York}
  \state{NY}
  \postcode{11201}
  \country{USA}
}

\author{Bei Yu}
\email{byu@cse.cuhk.edu.hk}
\author{Evangeline F.Y.~Young}
\email{fyyoung@cse.cuhk.edu.hk}
\affiliation{%
  \institution{Chinese University of Hong Kong}
%   \department{}
%   \streetaddress{}
  \city{Hong Kong}
%   \state{}
%   \postcode{}
%   \country{}
}

\author{Ramesh Karri}
\email{rkarri@nyu.edu}
% \affiliation{%
%   \institution{New York University}
%   \department{Center for Cybersecurity}
%   \streetaddress{2 Metrotech Center}
%   \state{NY}
%   \postcode{11201}

\author{Siddharth Garg}
\email{sg175@nyu.edu}

\affiliation{%
  \institution{New York University}
  \department{Center for Cybersecurity}
  \streetaddress{2 Metrotech Center}
  \city{New York}
  \state{NY}
  \postcode{11201}
  \country{USA}
}

\authorsaddresses{%
Authors' addresses: \\K. Liu (kang.liu@nyu.edu), B. Tan (benjamin.tan@nyu.edu), R. Karri (rkarri@nyu.edu), S. Garg (sg175@nyu.edu), Center for Cybersecurity, New York University.\\H. Yang (hyyang@cse.cuhk.edu.hk), Y. Ma (yzma@cse.cuhk.edu.hk), B. Yu (byu@cse.cuhk.edu.hk), E. F. Y. Young (\mbox{fyyoung@cse.cuhk.edu.hk}), Department of Computer Science and Engineering, Chinese University of Hong Kong}

%
% By default, the full list of authors will be used in the page headers. Often, this list is too long, and will overlap
% other information printed in the page headers. This command allows the author to define a more concise list
% of authors' names for this purpose.
\renewcommand{\shortauthors}{Liu et al.}

%
% The abstract is a short summary of the work to be presented in the article.

\begin{abstract}
%There are many challenging problems in electronic computer-aided design (CAD) flows, such as lithographic hotspot detection. 
%With recent advances in the realm of machine learning (ML) heralded by the introduction of deep learning techniques, 
There is substantial interest in the use of machine learning (ML) based techniques throughout the electronic computer-aided design (CAD) flow, particularly those based on deep learning. However, while deep learning methods have surpassed state-of-the-art performance in several applications, they have exhibited intrinsic susceptibility to adversarial perturbations --- small but deliberate alterations to the input of a neural network, precipitating incorrect predictions. In this paper, we seek to investigate whether adversarial perturbations pose risks to ML-based CAD tools, and if so, how these risks can be mitigated. To this end, we use a motivating case study of lithographic hotspot detection, for which convolutional neural networks (CNN) have shown great promise. In this context, we show the \emph{first} adversarial perturbation attacks on state-of-the-art CNN-based hotspot detectors; specifically, we show that small (on average 0.5\% modified area), functionality preserving and design-constraint satisfying changes to a layout can nonetheless trick a CNN-based hotspot detector into predicting the modified layout as hotspot free (with up to 99.7\% success). We propose an adversarial retraining strategy to improve the robustness of CNN-based hotspot detection and show that this strategy significantly improves robustness (by a factor of \textasciitilde3) against adversarial attacks without compromising classification accuracy.
%that does not add further computation during inference, nor compromise hotspot detection accuracy.
\end{abstract}



%
% Keywords. The author(s) should pick words that accurately describe the work being
% presented. Separate the keywords with commas.

%
% A "teaser" image appears between the author and affiliation information and the body 
% of the document, and typically spans the page. 
%%  \includegraphics[width=\textwidth]{sampleteaser}
%%  \caption{Seattle Mariners at Spring Training, 2010.}
%%  \Description{Enjoying the baseball game from the third-base seats. Ichiro Suzuki preparing to bat.}

%
% This command processes the author and affiliation and title information and builds
% the first part of the formatted document.
\maketitle

\section{Introduction}
% \todoblock{TODO: Need to work out what the standard terminology is... are we going to talk about Machine Learning/ML, Deep Learning (DL), Convolutional Neural Networks (CNN)...?}
% \todoblock{Introduce problem, talk lightly(?) about general trends re: Litho, NN etc. Add a pretty picture here to illustrate the design/feedback loop of design-simulate(verify)?}

Electronic system design flows provide several optimization and verification challenges as the scale and complexity of designs increases, placing a higher pressure on designers to deliver timely results.  There is substantial interest in using machine learning (ML) techniques for solving hard electronic computer-aided design (CAD) problems ranging from logic synthesis to physical design and design for manufacturability (DFM) \cite{moore_darpa_2018}.
A promised outcome of deep learning enhanced design flows is a faster and scalable development cycle, enabled by improvements in time-consuming steps of design space exploration \cite{greathouse_machine_2018}, logic optimization \cite{yu_developing_2018} and lithographic analysis \cite{yang_layout_2018}.

\begin{figure}[b]
\begin{center}$
\begin{array}{ccccc}

\text{horse}&\text{airplane}&\text{automobile}&\text{bird}&\text{cat}\\
\includegraphics[width=0.8in]{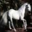}&
\includegraphics[width=0.8in]{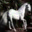}&
\includegraphics[width=0.8in]{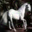}&
\includegraphics[width=0.8in]{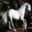}&
\includegraphics[width=0.8in]{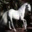}\\

% \text{deer}&\text{dog}&\text{fog}&\text{ship}&\text{truck}\\
% \includegraphics[width=0.8in]{Fig/Background/adv_4.png}&
% \includegraphics[width=0.8in]{Fig/Background/adv_5.png}&
% \includegraphics[width=0.8in]{Fig/Background/adv_6.png}&
% \includegraphics[width=0.8in]{Fig/Background/adv_8.png}&
% \includegraphics[width=0.8in]{Fig/Background/adv_9.png}\\

\end{array}$
\end{center}
\caption{A "clean" image of a horse (leftmost) and adversarial images with corresponding prediction labels. The adversarial perturbations are so minute as to appear imperceptible.}
\label{Fig:advperturb}
\end{figure}

Nonetheless, while deep learning methods have surpassed state-of-of-the-art performance on a wide range of applications, they have been shown to be brittle against adversarial perturbations \cite{GoodfellowSS14}.
Adversarial perturbations are small, imperceptible but targeted modifications to the input of the deep neural network, resulting in incorrect behavior. 
%\todoblock{%
For example, \autoref{Fig:advperturb} shows an image of a \texttt{horse} from the \texttt{CIFAR-10} dataset \cite{krizhevsky2009learning} --- each of the the subsequent four images are adversarially perturbed versions of the first that are classified as \texttt{airplane}, \texttt{automobile}, \texttt{bird} and \texttt{cat}, respectively. 
As noted earlier, the perturbations are so small that they are imperceptible.
%}  

Adversarial perturbations have been demonstrated in practically every application in which deep networks are used \cite{biggio_wild_2018}, 
and have raised fundamental questions about the ability of deep neural networks to generalize.
This leads to a natural question: \emph{what are the implications of adversarial perturbations on the security,  soundness, and robustness of deep learning techniques in CAD-related problems?}
While the CAD domain presents a challenge for adversaries, given the domain-specific knowledge required to perform stealthy (and meaningful) attacks, it is crucial to investigate whether adversarial perturbations pose a showstopping threat to this innovation.

As a motivating example, we study the
challenging CAD problem of lithographic layout hotspot detection. In physical design of an integrated circuit (IC), layout patterns are etched into silicon using optical lithography. Due to lithographic process variations, specific patterns are susceptible to manufacturing errors; these \textit{hotspots} need to be detected and fixed early in the IC design flow to avoid yield loss.
The conventional approach to hotspot detection is physics-based optical lithography simulations. While accurate, they are time consuming and computationally expensive for the full IC.
Noting that one can pose hotspot detection as image classification, recent work has proposed adoption of convolutional neural networks (CNN) for this problem, achieving state-of-the-art results~\cite{yang_layout_2018}.
% Not only is hotspot detection using CNNs accurate, but the query time of the CNN can be much faster than full simulation by a factor of \hl{Z}.
Once hotspots are detected, resolution enhancement techniques (RETs) such as optical proximity correction (OPC) and the insertion of sub-resolution assist features (SRAFs) can enhance IC layouts. Changes are verified using further lithography simulations, and iterated upon as required.

Now consider the following scenario where a designer is considering the purchase of a 3rd-party macro for their IC design. The designer wants to check the quality of the macro and has the IC layout images for verification. Using a CNN-based hotspot detector, the designer can quickly ascertain if the IC layout is printable as-is, and gauge the potential effort needed to correct any design flaws. To pass off a sub-par design as \textit{high quality}, the 3rd-party vendor selectively modifies the layout %\todoblock{Indicate that the modifications are minimal} 
to force the detector to misclassify hotspot regions as non-hotspot.
In other words, the attacker \textit{hides} hotspots by exploiting properties of the CNN --- identifying and taking advantage of the susceptibility of CNNs to adversarial perturbations.
However, malicious insertion is non-trivial. Unlike image perturbations that involve adding imperceptible noise \cite{biggio_wild_2018}, the attacker must add \textit{semantically meaningful} and realistic IC layout features to the design that pass design rule check (DRC), such as respecting spacing constraints. Successful attacks can have a significant impact: this sabotage can propagate undetected manufacturability issues, causing downstream reductions in IC yield and wasted designer effort.

% Now consider the following scenario, where a disgruntled in-house designer gains access to the GDSII files of the design.
% By selectively inserting patterns into the layout, such as additional sub-resolution assist features (SRAFs), the attacker can cause the CNN to misclassify a hotspot region as non-hotspot.
% This avenue for sabotage can result in the propagation of undetected manufacturability issues, causing downstream reductions in yield and subsequently wasted designer effort.
% Alternatively, the designer could insert patterns to increase the number of false-positives and therefore, the number of lithography simulations required, thus counteracting the productivity gain from adopting the CNN-based detector in the first place.

With lithographic hotspot detection as our motivating case study, we investigate, for the first time, a targeted attack on deep learning based CAD tools, to demonstrate the feasibility, challenges, and potential security implications for the CAD community. 
Our empirical findings highlight the need to study the underlying mechanics and to give wider consideration of the security and robustness implications of integrating ML-based tools into the IC design flow. Our contributions are threefold:

\begin{itemize}
    \item The first exploration of the impact of adversarial perturbations on deep neural network based CAD tools using IC lithographic hotspot detection as a case study.
    \item Comprehensive evaluation of two attack scenarios on CNN-based hotspot detectors: (1) white-box attacks, wherein the attacker has access to the model parameters of the detector and (2) black-box attacks, wherein the attacker has access only to the outputs of the detector.
    \item Exploration of adversarial retraining as a defense against adversarial perturbation attacks, yielding an equally accurate but robustified CNN for hotspot detection.
\end{itemize}

   The rest of the paper is organized as follows. 
   We explore the motivations for adopting deep learning in hotspot detection, outlining the motivations and goals of a potential attacker (\autoref{sec:mot}).
    This is followed by technical preliminaries to understand the principles of CNNs and the notion of adversarial perturbations (\autoref{sec:prelim}).
    In this work, our study centers around two CNN-based hotspot detectors and we detail the architecture and training of these detectors (\autoref{sec:case-study}).
    %These two CNN configurations will be targeted when we evaluate our proposed attacks to explore the efficacy of adversarial perturbations and to determine if there are any any differences in the natural attack resiliency between smaller CNNs and larger/deeper CNNs.
    Following this, we describe the attack methodologies, detailing how adversarial IC layouts can be generated to effectively \textit{hide} the presence of hotspots (\autoref{sec:attack-methods}).
    The first attack is a white-box attack, where the internals of the detector are available to the attacker.
    We then consider a more conservative attack, where the attacker can only query a black-box model.
    %To confirm our hypothesis that SRAF insertion in our attack does not universally fix hotspots, 
    We verify, via lithography simulations, that a vast majority of adversarially perturbed IC layouts are still hotspots but are not picked up as such by the hotspot detectors (\autoref{sec:attack-results}).
    Given the high success rate of our attack experiments, we propose robust retraining --- a promising defense against adversarial attacks --- presenting encouraging results (\autoref{sec:defense}).
    Our findings pose interesting questions that we then discuss (\autoref{sec:discussion}).
    To contextualize our work, we present related literature in deep learning for CAD and adversarial attacks (\autoref{sec:related}).
    %given that this work sits at the intersection of these two domains (\autoref{sec:related}).
    Ultimately, we conclude from this study that one should be aware of the limitations of using deep learning in CAD, and also be encouraged to investigate and adopt proactive countermeasures (\autoref{sec:conc}).

    % \todoblockgreen{Do we need to eventually add some numbers (results/percentages) here? (as signposts for the results to follow?)}

    % \todoblock{We present for the first time (as far as we know), a targeted attack on DL in EDA, and discuss the potential impacts/effects of doing this. We do not want to advocate that everything is bad, but to encourage diversity in approach, caution, etc.}
    % \todoblock{This paper is organized as follows.... (is there a classier way of outlining the paper?)}

\section{Motivation}
\label{sec:mot}

    %  \todoblock{To illustrate this threat, we discuss a popular/useful/contemporary application of DL in the EDA flow. First we explain what it is, and then we explain the potential threats}
     
\subsection{Deep Learning for Hotspot Detection}
    % \todoblockblue{CUHK: The main aim here is to say, okay, we have design pressures ... so we want things to be faster. This is how we make things faster (i.e., skip/avoid simulation, use lots of AI stuff), and something about trade-off in accuracy/speed. Headings are suggestions only.}
    
   %%BY HAOYU

    % \todoblock{Details about the Lithography problem, including some background on SRAF insertion, hotspots etc.}
    % \begin{itemize}
    %     \item SRAFs
    %     \item Litho-process, hotspot definitions
    %     \item GDSII format/layout image
    % \end{itemize}
\subsubsection{Lithographic Hotspot Detection}
% \todoblock{Should it be \emph{lithographic} hotspot detection?}
In advanced technology nodes the layout feature sizes are much smaller than the light wavelengths used in the optical lithography systems. As a result, complex interactions between light patterns in lithography have made printed patterns sensitive to process variations. This has increased challenges in IC back-end design and sign-off flows. Lithography induces defects due to phenomena such as diffraction, resulting in lithographic \emph{hotspots} \cite{yang_layout_2018,yang2019hotspot,reddy_enhanced_2018}.

\begin{figure}[t]
        \centering
        \subfigure[Layout with vias only]{\label{fig:layout_noret}\includegraphics[height=1.2in]{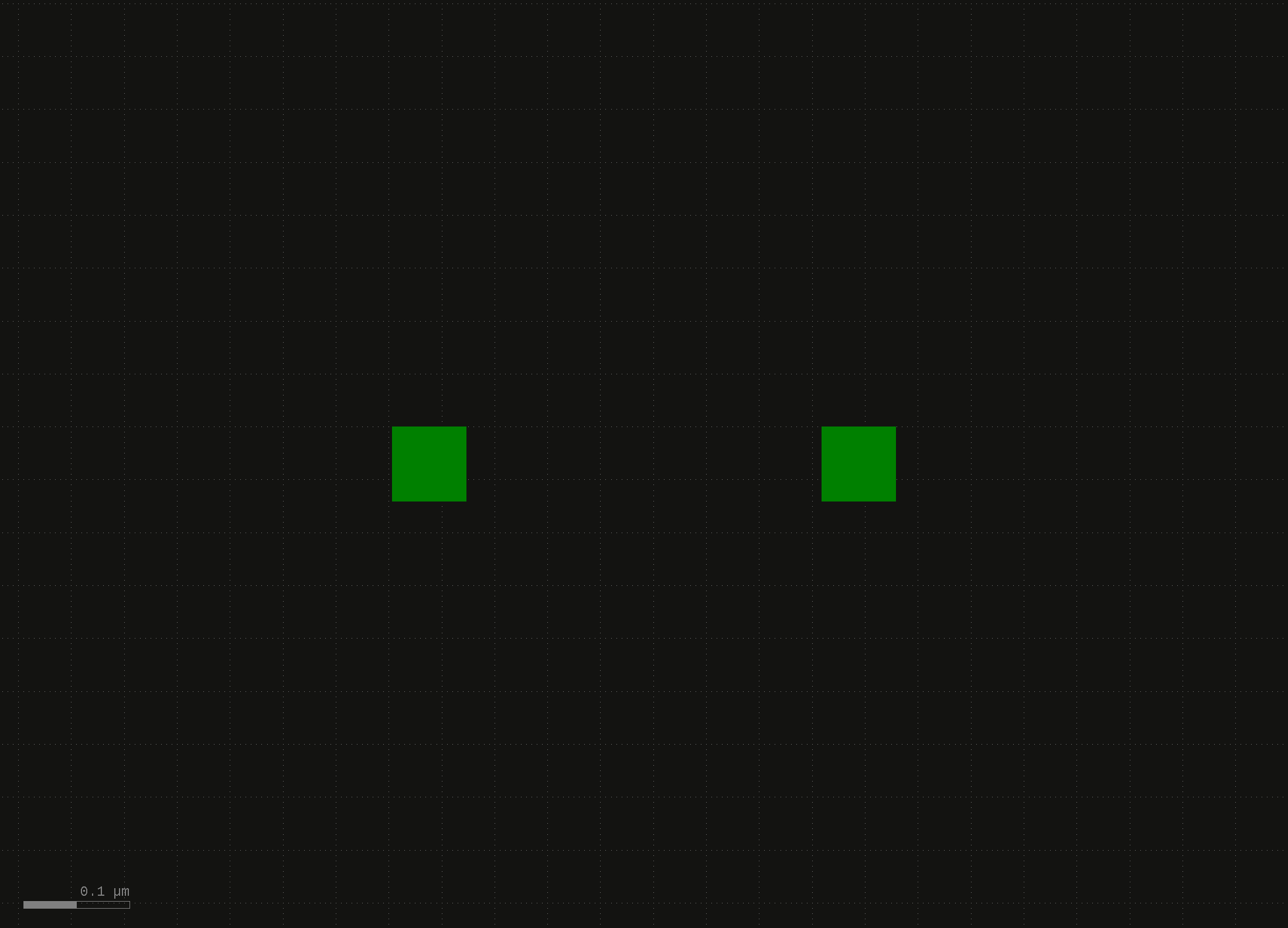}}
        \hspace{1.5em}
        \subfigure[Lithography simulation of layout with vias only]{\label{fig:litho_noret}\includegraphics[height=1.2in]{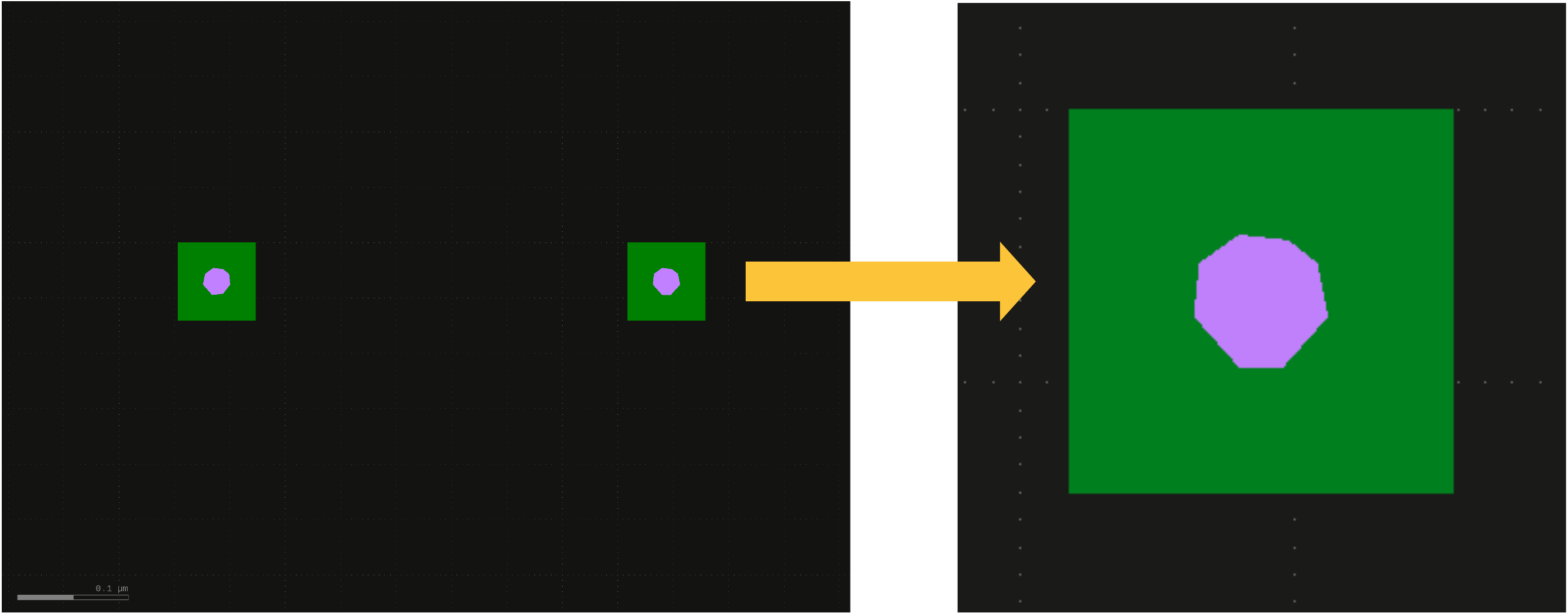}}
        \subfigure[Layout with vias and SRAFs]{\label{fig:layout_ret}\includegraphics[height=1.2in]{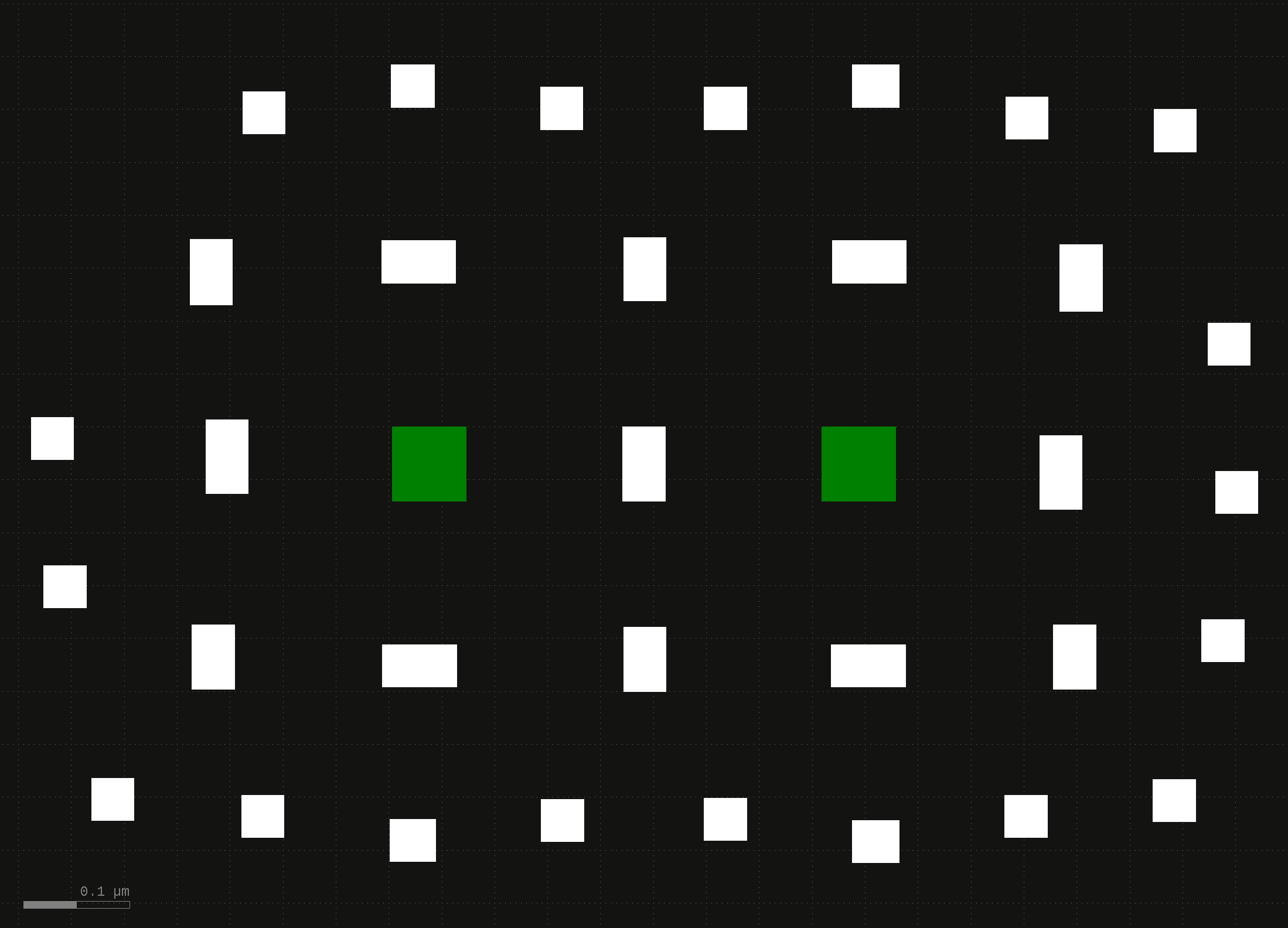}}
        \hspace{1.5em}
        \subfigure[Lithography simulation of layout with vias and SRAFs]{\label{fig:litho_ret}\includegraphics[height=1.2in]{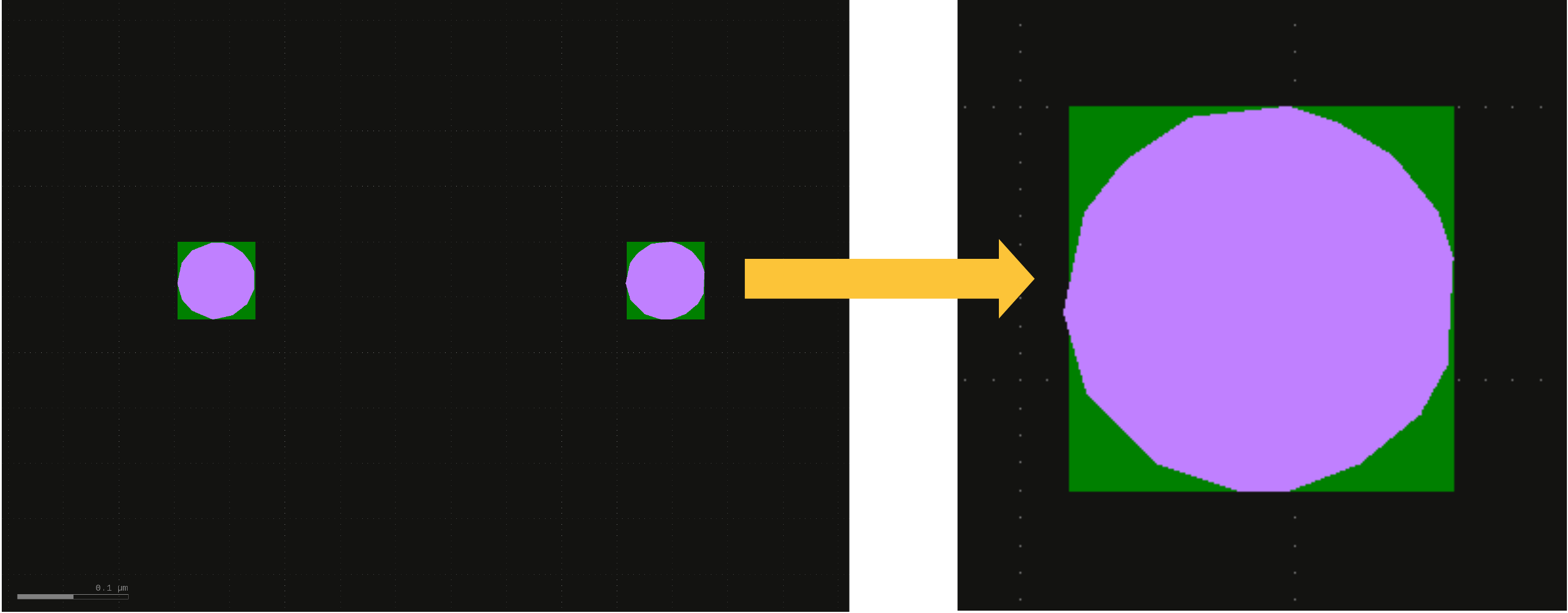}}
        \caption{Illustration of lithography simulation results of layouts with vias only, and both vias and SRAFs.}
        \label{fig:ret}
    \end{figure}

% Due to the mismatch between the lithography system and manufacturing technology node,
% we are facing challenges in IC backend design and sign-off flows.
% \todoblock{Not clear what ``mismatch" refers to? Clarify...wavelength of light greater than the minimum feature size?}
% One of the challenges is defects induced in the during lithography (known as lithography hotspots) , caused by the diffraction information loss in the lithography procedure.
% \todoblock{``diffraction information loss" not clear.}
Consider \autoref{fig:ret}(a), which shows an IC layout containing two vias colored green. If this layout were printed as is, the resulting printed output would be unsatisfactory. Only a small region of the desired vias is printed --- shown in purple in \autoref{fig:ret}(b). Thus, resolution enhancement techniques (RETs) such as sub-resolution assist features (SRAF) \cite{geng2019sraf} and optical proximity correction (OPC) \cite{yang_gan-opc:_2018} have been proposed to ease IC layout manufacturability; they aim to compensate for distortion during lithography. \autoref{fig:ret}(c) shows the effect of SRAF insertion. The printed pattern more accurately reflects the required pattern (\autoref{fig:ret}(d)). However, even when equipped with rigorous RETs, the layout can have hotspots due to unpredictable lithography process variations. Therefore, it is vital to spot potential hotspots before manufacturing and correct them either by using RET or by re-design.
%Traditional methods rely highly on time-consuming and precise lithography simulation.

\subsubsection{Deep Learning Based Hotspot Detection}
    % \todoblock{Something about speed-up, accuracy gains?}
In light of the prohibitive run-time of lithographic simulation, recent work has sought to speed-up hotspot detection using pattern matching \cite{yu2012accurate} and machine learning \cite{matsunawa2015new,yang_layout_2017}.
Pattern matching methods find similar or identical hotspot-causing patterns in a new design from a library of known hotspots.
% is accurate and fast when selecting already known hotspots.
These techniques are both fast and accurate if the patterns are similar to those in the library, but cannot find previously unseen hotspot patterns.
In contrast, machine learning solutions seek to capture the underlying \emph{physics} of lithographic simulation (i.e.,
the relationships between IC layout features and 
their manufacturability) and, as such, \emph{generalize} to unseen patterns (or at least that has been the hope). 
%
%are expected to perform accurate hotspot detection efficiently after expending the effort needed for good feature engineering.
Recent advancements on CNN based hotspot detection~\cite{yang_layout_2018,yang2019hotspot} 
%provide alternatives for efficient hotspot detection thanks to its automatically learned features and GPU-friendly computation.
%For example, state-of-the-art works \cite{yang_layout_2018,yang2019hotspot} 
have shown that both shallow and deep CNNs are more accurate compared to legacy machine learning based and pattern matching based techniques. 
%several metrics from detection accuracy to efficiency.
% \todoblock{``Efficiency" undefined. Speed? Computational cost? --- BT: Speed I think, although I Saw that mentioned in LithoGAN to refer to simulation speed up rather than HS detection. We can leave this out.}
%It is this promise of design flow improvement from deep learning that motivates our examination of potential threats from adversarial perturbations.

%   \begin{figure}[b]
%     \begin{center}$
%     \begin{array}{cc}
%     \includegraphics[width=2.2in]{Fig/Experiment/mask-9999.png}&
%     \includegraphics[width=2.2in]{Fig/Experiment/nominal-9999.png}    \end{array}$
%     \end{center}
%     \caption{Examples on input layout and the corresponding printed pattern.}
%     \label{fig:litho-ex}
%     \end{figure}
    
    % \todoblockgreen{BT: Could we use this \autoref{fig:litho-ex} to show desired layout without RET | output without RET | layout with RET | output with RET?}
    
    % \begin{figure}[b]
    % \begin{center}$
    % \begin{array}{cc}
    % \includegraphics[width=1.5in]{Fig/Background/mask-noret.png}&
    % \includegraphics[width=1.5in]{Fig/Background/nominal-noret.png}&
    % \includegraphics[width=1.5in]{Fig/Background/nominal-noret-zoom.png}\\
    % \includegraphics[width=1.5in]{Fig/Background/mask-ret.png}&
    % \includegraphics[width=1.5in]{Fig/Background/nominal-ret.png}&
    % \includegraphics[width=1.5in]{Fig/Background/nominal-ret-zoom.png}
    % \end{array}$
    % \end{center}
    % \caption{Examples on input layout with/without RET and the corresponding printed pattern.}
    % \label{fig:litho-ex}
    % \end{figure}
   
\subsection{Threat Model}
    \subsubsection{Setting}
        To motivate our work in examining the security and robustness of deep learning in CAD, we explore the scenario of a designer considering the purchase of a macro from a 3rd party IP vendor.
        The 3rd party IP vendor distributes hard macros in GDSII format \cite{calma_company_gdsii_1987}, where the circuit is laid out and allegedly enhanced for lithography using RETs.
        As part of the validation process, the designer does a "sanity-check" on the macro to establish its quality by using a CNN-based hotspot detector (which may be a commercial tool in a local or cloud setting).
    
    \subsubsection{Attack Goals}
        The vendor aims to sell low-quality hard macros, either to make a profit from their design short-cuts or to sabotage the designer (by forcing them to waste time and resources in rectifying poor designs).
        To achieve this aim, the attacker's goal is, therefore, \emph{to fool the target CNN-based hardware detector into classifying hotspots as non-hotspot.}
        This should be achieved by making the smallest changes to the layouts as possible.
        In this work, we investigate SRAF insertion as an RET; consequently, the aim of the attacker is to insert as few SRAFs as possible.
        
    \subsubsection{Attacker Capabilities}
        In the context of deep learning, the attacker capabilities can be defined based on the amount of information they possess about the network under attack. 
        This includes information about the network's hyper-parameters: its overall architecture, its weights and biases, the training algorithms and training data, etc. 
        For this case study, we consider two scenarios: (1) an attacker with white-box access, where they have full knowledge of the CNN, including its network architecture, weights and biases; 
        and (2) an attacker with only black-box access, where they are able to query the detector, receiving both output classification as well as the accompanying prediction confidence. 
        Both models have been studied in prior work~\cite{GoodfellowSS14,papernot2017practical,delving,biggio_wild_2018}. % and reflect real-world settings. 

\section{Deep Learning Preliminaries}
\label{sec:prelim}

To appreciate the potential of deep learning for CAD problems such as hotspot detection, we present relevant technical preliminaries for CNNs and adversarial perturbations.

\subsection{CNN Basics}
    % \todoblock{What is a CNN, how is it used...}
    A CNN features an input layer, a number of hidden layers, and an output layer.
    The CNN takes in some input (e.g., an image) and propagates the data through a series of linear and non-linear operations (akin to convolution and activation of "neurons").
    After all the input has been transformed by each of the hidden layers, the final output produces a classification prediction for the input.
    The CNN is "trained" by configuring the parameters of the filters in each layer (the weights).
    
    We can express this formally as follows.
    A neural network is defined as a function $F$ that takes input $x\in \mathbb{R}^N$ and gives output $z\in \mathbb{R}^m$, such that $z=F(x)$.  For an $m$-class classifier, we define $z$ as an array, $z=[z_1,z_2,\cdots,z_m]$, where $z_i$ is the prediction probability of class $i$, $i=1,2,\cdots,m$.
    The network output $z$ is subject to the constraints: $0\le z_i \le 1$, and $z_1+z_2+\cdots+z_m=1$. The label $y$ of input $x$ takes the output class with the highest prediction probability, such that $y=\text{label}(x)=\arg\max_i z=\arg\max_i F(x)$. A deep neural network classifier has multiple layers of neurons, the last being a softmax layer. Hence, the neural network can be expressed as:
        \begin{equation}
            F = \text{input layer} \circ F_1 \circ F_2 \circ \cdots \circ F_{k-1} \circ F_k \circ \text{softmax}
        \end{equation}
    where
        \begin{equation}
            F_i(x) = f_i(w_i x+b_i), \quad i = 1,2,\cdots,k
        \end{equation}
    Here $f_i$ is the activation function of layer $F_i$, and $w_i$ is the model weights and $b_i$ is the bias. Some common choices of activation function $f$ include logistic, tanh and ReLU \cite{nair_rectified_2010}.
    In an image classification neural network, input $x$ is either a grey-scale image with one channel or an RGB image with three channels, where each channel of pixel $x_i$ takes integer values from  $[0,255]$. %Sometimes, these pixel values are scaled within $[0,1]$ for .

\subsection{Adversarial Perturbations}
    The existence of adversarial inputs for classification using neural networks was first described by Szegedy et al. \cite{szegedy_intriguing_2013}. They observed a phenomenon whereby neural networks would change its output prediction based on imperceptible perturbations in the input.
    In these cases, while the network would be "fooled", a human would not be "fooled".  This property can be exploited by an adversary, whereby inputs can be crafted to fool a target network and cause misclassification.
    
    Formally, let $y^*$ denote the true label of a clean input $x$, and $y$ denote the prediction label of $x$ given by the neural network. 
    The adversary aims to generate an adversarial input $x'$ close to $x$, and mislead the network to output a target label $y'$, while $y' \ne y$.
    The difference between $x$ and $x'$ is measured by a distance metric and constrained by a constant $\delta$, such that $\|x - x^\prime\| \leq \delta$.
    Normally $\delta$ is so small to be perceptual to human eyes and should not change the prediction label from $y$ to $y'$.

    In non-targeted attacks, adversaries search for adversarial inputs $x'$ as long as its output label $y' \ne y$. In targeted attacks, the target label is pre-defined by the adversary, and $y'$ could be quite distinct than $y$.
    There are several schemes for crafting adversarial perturbations.
    Our work is inspired by the following methods that have been explored in a general adversarial perturbation context.

% \subsection{Fast Gradient Sign Methods}
    \subsubsection{Basic fast gradient sign (FGS) method}
    Goodfellow et al. \cite{GoodfellowSS14} proposed the FGS method for adversarial input generation.  For non-targeted attacks, starting with a clean input $x$, the adversary moves each pixel in the opposite direction of the gradient of the loss function of the true label with respect to $x$.
    The goal is to mislead the network into outputting any label other than the true label. 
    The non-targeted FGS attack can be described mathematically as follows:
    %Equation \ref{Equ:fgs_ntg}. 
    \begin{equation}\label{Equ:fgs_ntg}
        x^\prime \gets \text{clip}(x + \epsilon\text{sign}(\nabla\ell_{F,y^*}(x))).
    \end{equation}
    $\epsilon$ is a small constraint scalar, $\ell$ is the loss function and $\text{clip}(x)$ ensures pixel values fall in the desired range.
    
    On the other hand, in a targeted attack, the adversary seeks to fool the network into misclassifying $x$ as a specific target label. This is achieved
    %increases the prediction probability of a target label 
    by altering pixels in the direction of the gradient of the loss function of the target label with respect to $x$.  The attack is described by \autoref{Equ:fgs_tg}.
    \begin{equation}\label{Equ:fgs_tg}
        x^\prime \gets \text{clip}(x - \epsilon\text{sign}(\nabla\ell_{F,y'}(x)))
    \end{equation}
    
    %Be notified that FGS method is essentially a $L_1$ metric based attack. It is natural to consider other $L_p$ metrics and come up with similar attacks. But different attacks could differ in the success rate when constrained by the same distortion amount. Apparently basic FGS methods are designed to be fast rather than search for adversarial inputs with the minimum distortion. In FGS most pixels in the image would be altered.
    These two attacks emphasize computational efficiency and speed at the expense of introducing relatively large perturbations. Sophisticated techniques that seek to find the smallest possible perturbation, albeit at greater computational expense, have subsequently been proposed~\cite{kurakin_adversarial_2016} --- one such attack is described next. 

\begin{comment}
It is natural to consider the $L_2$ metric and come up with a similar fast gradient value (FGV) attack \cite{liu2016delving}. The non-targeted and targeted FGV methods are shown by Equation \ref{Equ:fgv_ntg} and Equation \ref{Equ:fgv_tg}, respectively.
\begin{equation}\label{Equ:fgv_ntg}
    x^\prime \gets \text{clip}\bigg(x + \epsilon\frac{\nabla\ell_{F,y^*}(x)}{\|\nabla\ell_{F,y^*}(x)\|_2}\bigg)
\end{equation}
\begin{equation}\label{Equ:fgv_tg}
    x' \gets \text{clip}\bigg(x - \epsilon\frac{\nabla\ell_{F,y'}(x)}{\|\nabla\ell_{F,y'}(x)\|_2}\bigg)
\end{equation}
\end{comment}

\subsubsection{Iterative fast gradient sign (IFGS) methods}
IFGS methods operate over multiple iterations, adding relatively small perturbations in each  \cite{kurakin_adversarial_2016}.
%In each iteration, the same as in the basic FGS method, an adversary moves each pixel along a certain direction based on the gradient of loss function respect to $x$, scaled by a constant $\alpha$, and followed by the clip function. After executing multiple and limited steps of the basic FGS method, the adversary manipulate the clean inputs into adversarial ones for non-targeted or targeted attacks. The pixels of final adversarial inputs are still constrained by the constant scalar $\epsilon$ as in basic FGS methods.
%IFGS methods for non-targeted and targeted attacks are shown in Equations \ref{Equ:ifgs_ntg} and \ref{Equ:ifgs_tg}, respectively. 
As such, IFGS methods can generate adversarial inputs with smaller distortion when compared to basic FGS. Equation~\ref{Equ:ifgs_ntg} and Equation~\ref{Equ:ifgs_tg} describe the updates performed by the non-targeted and targeted versions of IFGS in each iteration. As an example, the adversarial perturbations in Fig. \ref{Fig:advperturb} were generated by the IFGS method.

\begin{equation}\label{Equ:ifgs_ntg}
    x'_0 = x, \quad x_{N+1}^\prime \gets \text{clip}_\epsilon(x'_N + \alpha\text{sign}(\nabla\ell_{F,y^*}(x)))
\end{equation}
\begin{equation}\label{Equ:ifgs_tg}
    x'_0 = x, \quad x_{N+1}^\prime \gets \text{clip}_\epsilon(x'_N - \alpha\text{sign}(\nabla\ell_{F,y'}(x)))
\end{equation}

\subsubsection{Semantically meaningful perturbations}
 Another body of work has focused on semantically meaningful perturbations. For instance, specially crafted stickers affixed to traffic signs can mislead traffic sign classifiers~\cite{Eykholt_2018_CVPR}. These perturbations are not imperceptible, in fact, quite the opposite, they are easily spotted, but are designed to seem innocuous. %Yet, they are semantically meaningful in that 
 For instance, a human is unlikely to think that a small sticker on a traffic sign indicates an adversarial attack. Our work crafts such perceptible but semantically meaningful perturbations. However, the notion of what is semantically meaningful is informed by the underlying domain of lithography.

\section{Case Study: IC Lithographic Hotspot Detection}
\label{sec:case-study}
    In this work we use two different CNN-based hotspot detectors to explore our proposed attacks. % on representative large and small CNNs.
    They are trained using the same dataset and act as \textit{targets} for adversarial perturbations.
    This section describes details of our dataset, the network architectures, and the training process.
    Our case study draws heavily from prior work \cite{yang_layout_2018}.
    
    \subsection{Layout Dataset}
        \label{sec:dataset}
        Existing datasets for lithographic hotspot detection, for instance, the widely used ICCAD '12 contest dataset~\cite{iccad-dataset}, do not come with much of the information required to verify the success of adversarial attacks. For instance, the ICCAD '12 data specifies neither design rules nor does it specify lithography simulation parameters.   
        Therefore, for this case study, we prepared our own layout dataset comprising of 10403 layout clips stored in the GDSII format. We targeted the detection of lithographic hotspots for via layers using SRAF-based RET.
        To create the large number of layout samples, we generated the via patterns in the following manner: 
        \begin{enumerate}
            \item Within each clip region ($2~\mu m \times 2~\mu m$) we place lower layer metal gratings with fixed wire critical dimension (CD) and pitch;
            \item We add an upper metal layer with preset CD and spacing constraints;
            \item The cross regions between two metal layers become candidates for via placement --- we place vias stochastically with a given probability;
            \item Finally, vias that violate design rules are removed. In this dataset, we use vias sized 70~nm $\times$ 70~nm and enforce a minimum via spacing of 70~nm.
        \end{enumerate}
        
         \begin{figure}[b]
            \centering
            \subfigure[Example Layout]{\label{fig:via_sraf_ban}\includegraphics[height=1.3in]{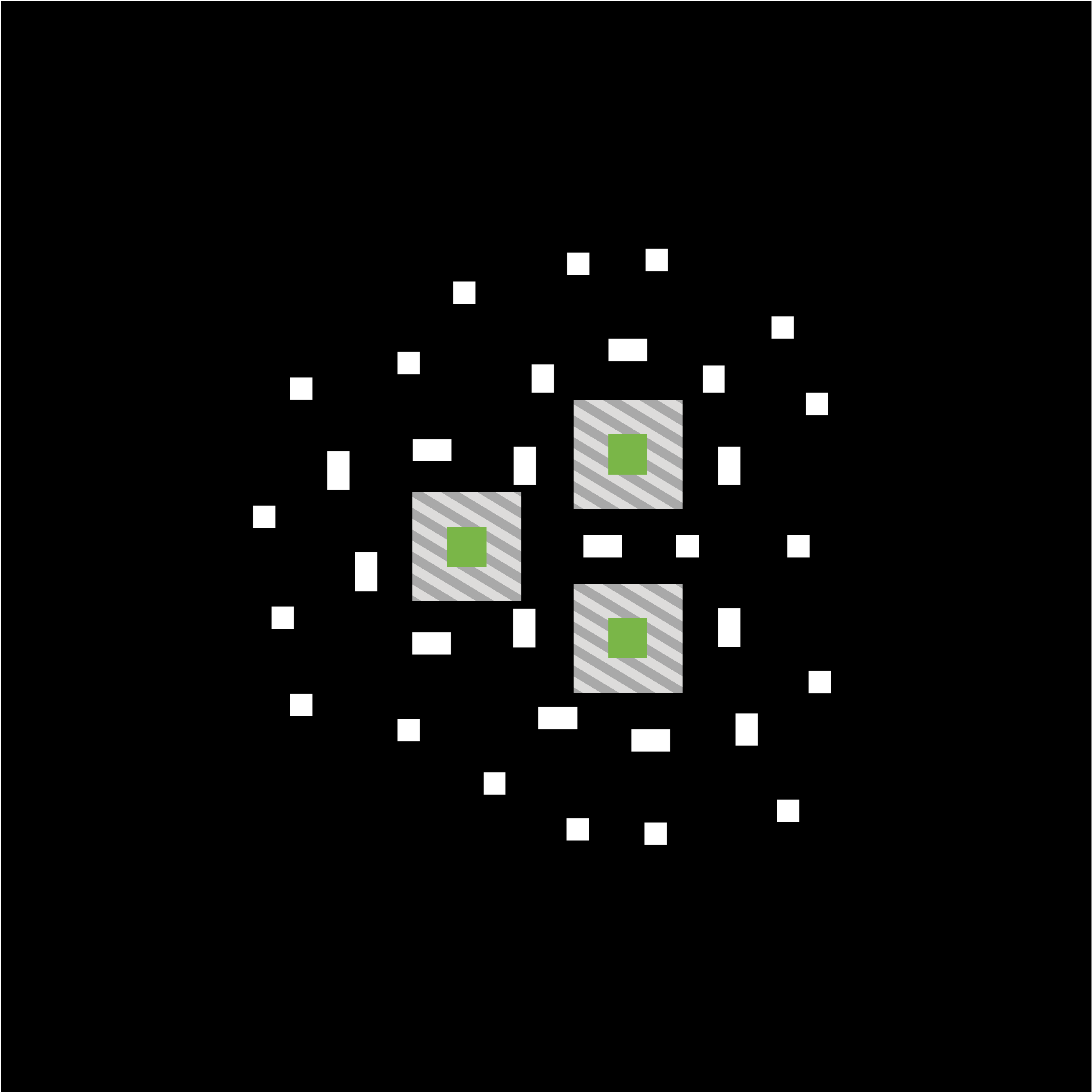}}
            \hspace{3em}            
            \subfigure[Edge Placement Error (EPE)]{\label{fig:EPE-ex}\includegraphics[height=1.3in]{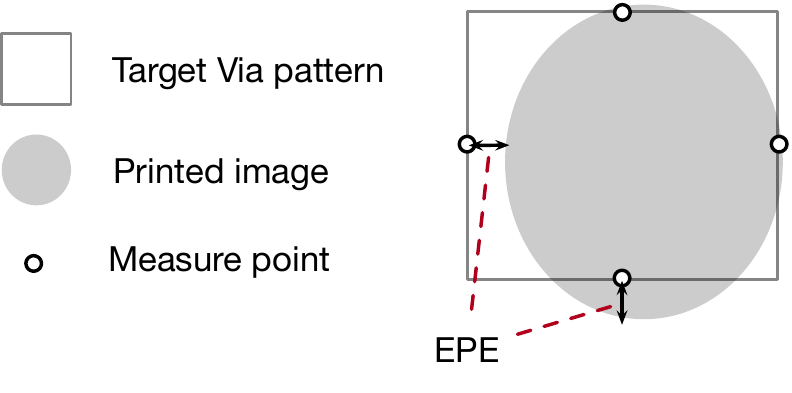}}
            \caption{(a) Example of a layout with vias (in green), SRAFs (in white), and the forbidden areas (striped region). (b) Illustration of edge placement error (EPE). Each target via pattern has four measure points (one at the center of each edge). The EPE is the perpendicular displacement from the measure point to the corresponding printed image (contour).}
            \label{fig:epe_via_sraf_ban}
        \end{figure}
        
        Once the "raw" layouts are produced, we perform optical proximity correction (OPC) and lithography simulation using Mentor Calibre \cite{mentor_graphics_calibre_nodate} to insert SRAFs; we set the allowable SRAF region to a 100~nm~--~500~nm city-block distance.
        An example of a layout clip is shown in \autoref{fig:via_sraf_ban}.
        %\textcolor{red}{An example of layout clip is shown in fig 3a.}
        Next, we determine the ground truth hotspot/non-hotspot labels for the layouts.
        In this work, we use the edge placement error (EPE) as our metric for determining the quality of the printed patterns.
        Each via pattern in a layout is associated with four measure points, with one point at the center of each edge. 
        The EPE is defined as the perpendicular displacement from the measure point to the corresponding printed contour, as illustrated in \autoref{fig:EPE-ex}. 
        A layout is identified as a hotspot layout if there exists any measure points with the EPE greater than 2~nm, as in typical industrial settings.
        % I think this is a user/process defined size ... will confirm with CUHK, but there are a few works that mention EPE, and one that mentions 3~nm for 20nm process? Something about a spec limit? LithoGAN -> https://www.cerc.utexas.edu/utda/publications/C191.pdf -> http://citeseerx.ist.psu.edu/viewdoc/download?doi=10.1.1.821.3435&rep=rep1&type=pdf

        As shown in \autoref{fig:via_sraf_ban}, the layouts we produced have three key features: vias (the desired pattern to be printed), SRAFs (used to improve printability), and forbidden regions (where SRAFs should not be placed).
        Each via is surrounded by a square forbidden region whose edges are 100~nm from the via's edges.
        The GDSII files contain three layers of interest: (1) a via layer, (2) an SRAF layer, and (3) a forbidden region layer.
    
        \subsection{Design of CNN-based hotspot detectors}
        \label{sec:cnn-design}
        
        % \subsubsection{Hotspot Detection Networks}
        % To examine how our attack methodology is affected by network architecture and baseline hotspot detection accuracy, we first trained two different CNN-based hotspot detectors to see if our attacks could be successful:%, as well as to test two theories that we had:
        Using the layout dataset we trained two different CNN-based hotspot detectors to represent networks of different complexity, adopting procedures described in prior work \cite{yang_layout_2018}. The parameters are shown in \autoref{tab:network-architecture} and \autoref{tab:network-architecture-bc}.  % to see if our attacks could be successful:
        \begin{itemize}
            \item \textit{Network A} is a smaller 9-layer network that is fast(er) to train and for prediction. We observed that further increasing network depth/complexity did not increase accuracy; i.e., Network A is "right-sized" for accuracy. 
            \item \textit{Network B} is a larger 15-layer network that is
            %has more layers than Network A.
            slower to train, but is potentially  
            %hypothesize that it 
            less susceptible to attack as the complex architecture learns sophisticated features for hotspot detection. Prior work on adversarial robustness suggests that deeper, complex network are more resilient to attack~\cite{madry_towards_2017}. 
            
        \end{itemize}
        %SG: above makes sense? The goal is to justify any design choices.
        \subsubsection{Data Preprocessing}
        The dimension of the GDSII layouts is 2000~nm $\times$ 2000~nm, which can be represented as 2000 pixel $\times$ 2000 pixel binary-valued images, where all the layers are flattened.
        Layout polygons are represented with pixel intensity of 255 and the background is represented with a pixel intensity of 0.  For training and inference we scale the layout image by a factor of 255 so that all the pixel intensities are either 1 or 0.
       
        Training a CNN on large images requires significant computation resources and time. Therefore, as proposed in~\cite{yang_layout_2018},
        % (\textcolor{red}{lets please use citations or have justifications for any design choices})
        we compute a discrete cosine transformation (DCT) on each image to extract its features as input for the networks. The equation for DCT is shown in Equation~\ref{eqn:dct}. 
        
        \begin{equation}
        \label{eqn:dct}
        D_{k_1,k_2} = \sum_{n_1=0}^{N_1-1}\sum_{n_2=0}^{N_2-1}I_{n_1,n_2}\cos{\bigg[\frac{\pi}{N_1}\bigg(n_1 + \frac{1}{2}\bigg)k_1\bigg]}\cos{\bigg[\frac{\pi}{N_2}\bigg(n_2 + \frac{1}{2}\bigg)k_2\bigg]}
        \end{equation}
        
        Here $n_1$ and $n_2$ are the horizontal and vertical coordinates of the image pixels, and $N_1$ and $N_2$ are the width and height of the image. $k_1$ and $k_2$ represent the horizontal and vertical coordinates of the DCT coefficients. To reduce the image dimensions, we perform DCT on non-overlapping 100 pixel $\times$ 100 pixel sub-blocks on each layout image (with a 100 pixel stride), and then keep a selection of DCT coefficients. For Network A, we keep the coefficients of the 32 lowest frequencies, producing inputs of size (20, 20, 32).
        For Network B, we keep the coefficients of the 36 lowest frequencies (i.e., more information for the larger network), producing inputs of size (20, 20, 36).
        This speeds up training with low loss of information without affecting network performance.
        
        \begin{table}[t]
        \centering
        \caption{Architecture of Network A.}
        \label{tab:network-architecture}
        \begin{tabular}{@{}llll@{}}
        \toprule
        Layer & Kernel Size & Stride & Output Size \\ \midrule
         input          &   -   &   -   &   (20, 20, 32)    \\
         conv1\_1       &   3   &   1   &   (20, 20, 16)    \\
         conv1\_2       &   3   &   1   &   (20, 20, 16)    \\
         maxpooling1    &   2   &   2   &   (10 , 10, 16)   \\
         conv2\_1       &   3   &   1   &   (10, 10, 32)    \\
         conv2\_2       &   3   &   1   &   (10, 10, 32)    \\
         maxpooling2    &   2   &   2   &   (5, 5, 32)      \\
         fc1            &   -   &   -   &   250             \\ 
         fc2            &   -   &   -   &   2               \\ \bottomrule
        \end{tabular}
        \end{table}

        \begin{table}[t]
        \centering
        \caption{Architecture of Network B.}
        \label{tab:network-architecture-bc}
        \begin{tabular}{@{}llll@{}}
        \toprule
        Layer & Kernel Size & Stride & Output Size \\ \midrule
         input          &   -   &   -   &   (20, 20, 36)    \\
         conv1\_1       &   3   &   1   &   (20, 20, 16)    \\
         conv1\_2       &   3   &   1   &   (20, 20, 16)    \\
         conv1\_3       &   3   &   1   &   (20, 20, 16)    \\
         maxpooling1    &   2   &   2   &   (10, 10, 16)    \\
         conv2\_1       &   3   &   1   &   (10, 10, 32)    \\
         conv2\_2       &   3   &   1   &   (10, 10, 32)    \\
         conv2\_3       &   3   &   1   &   (10, 10, 32)    \\
         maxpooling2    &   2   &   2   &   (5, 5, 32)      \\
         conv3\_1       &   3   &   1   &   (5, 5, 64)      \\
         conv3\_2       &   3   &   1   &   (5, 5, 64)      \\
         conv3\_3       &   3   &   1   &   (5, 5, 64)      \\
         maxpooling3    &   2   &   2   &   (3, 3, 64)      \\
         fc1            &   -   &   -   &   500             \\ 
         fc2            &   -   &   -   &   2               \\ \bottomrule
        \end{tabular}
        \end{table}
        
        \subsubsection{Training}
        We train both networks using the same layout dataset.
        We randomly split 10403 layout images into 8000 training images and 2403 test images, where the training data consists of 2774 hotspot and 5226 non-hotspot images, and test data has 841 hotspot and 1562 non-hotspot images.
        To compensate for data imbalance, we incorporate class weights to weigh the loss function during training, which tells the model to "pay more attention" to samples from an under-represented class~\cite{he2008learning}.
        This is done for both networks to achieve a balanced hotspot and non-hotspot detection accuracy.
        We implement network training with the Keras library \cite{chollet2015keras}, and use the ADAM optimizer \cite{kingma_adam:_2014} for loss minimization. 
        The confusion matrix is shown in \autoref{tab:confusion-matrix}.
        Our training of the baseline networks follows the same methodology as in prior work~\cite{yang_layout_2018}. Although Network A and Network B have the same overall\footnote{Overall accuracy is defined as the average of non-hotspot classification accuracy and hotspot classification accuracy.} and hotspot prediction accuracy, we seek to explore the robustness of both networks with different depths/complexity.
        
        \begin{table}[t]
        \centering
        \caption{Confusion matrix of networks A and B.}
        \label{tab:confusion-matrix}
        \begin{tabular}{clcccc}
        \hline
        \multicolumn{2}{c}{\multirow{3}{*}{}} & \multicolumn{4}{c}{Prediction} \\ %\cline{3-6} 
        \multicolumn{2}{c}{} & \multicolumn{2}{c}{Network A} & \multicolumn{2}{c}{Network B} \\ \cline{3-6} 
        \multicolumn{2}{c}{} & non-hotspot & hotspot & non-hotspot & hotspot \\ \hline
        \multicolumn{1}{l}{\multirow{2}{*}{Condition}} & non-hotspot & 0.72 & 0.28 & 0.72 & 0.28 \\
        \multicolumn{1}{l}{} & hotspot & 0.29 & 0.71 & 0.28 & 0.72 \\ \hline
        \end{tabular}
        \end{table}

\section{Proposed Attack Methodologies}
\label{sec:attack-methods}
\subsection{Overview}
    We propose attack methodologies for modifying layouts with hotspots such that they fool the CNN-based hotspot detector into misclassifing layouts as non-hotspot.
    We experiment with two attack types: a white-box attack, where the attacker has full access to the internal details (weights, architecture, etc.) of the hotspot detector, and a black-box attack, where the attacker can only query the detector to receive the output prediction and associated confidence.
    % \subsubsection{Attack Constraints}
    During the attack, the attacker aims to fool the target detector by modifying layouts in a \textit{semantically meaningful} way.
    This means that the attacker cannot alter the IC layout by moving via locations as this may change design functionality; in this attack we only add SRAFs to the layout.
    Further, the modifications must be small and innocuous, for instance, by only using shapes that already exist in the layout dataset.
    Finally, the perturbations should not introduce DRC violations.
    Based on these considerations, our perturbations must satisfy the following constraints:
 %   \subsubsection*{Attack Constraints}
    \begin{enumerate}
        \item Insertion Constraint: Maliciously-inserted SRAFs can only be added to the SRAF layer.
        \item Shape Constraint: Maliciously-inserted SRAFs should be rectangles, with a fixed width of 40~nm. The height can be selected within 40~nm -- 90~nm, at a resolution of 1~nm. The SRAF can be placed either horizontally or vertically.
        \item Spacing Constraint: The Euclidean distance between any two SRAFs should be at least 40~nm. 
        \item Forbidden Zone Constraint: Maliciously-inserted SRAFs cannot overlap with the forbidden region in a layout.% (as represented in layer 21/0).
    \end{enumerate}
    For simplicity, the our attack evaluation involves adding 40~nm wide SRAFs with the following height options: 40, 50, 60, 70, 80 or 90~nm, all placed horizontally.  
    
    \subsection{White-box Attack}% -- a gradient-guided approach}
    In the white-box attack, the attacker knows the internal details of the target CNN based hotspot detector, and exploits this as part of the attack.  We propose a gradient-guided approach to generate adversarial layouts, inspired by the fast gradient sign approach \cite{GoodfellowSS14} (explained in \autoref{sec:prelim}). 
    Since the baseline hotspot detection networks (in \autoref{sec:cnn-design}) take DCT coefficients as inputs, a na{\"i}ve attack would need to modify these coefficients and then perform inverse-DCT to produce adversarially perturbed layouts. There are at least three reasons why this na{\"i}ve approach is infeasible:
    
    \begin{enumerate}
        \item There is not enough data to reconstruct any layout without information loss, since the input DCT coefficients used for inference are only the low frequency components.
        \item There is no guarantee that modifications of DCT coefficients, when reflected back to layout images, satisfy the attack constraints above. %, such as the shape rule, spacing rule, and forbidden zone rule.
        \item It is challenging to modify DCT coefficients that result in an exact 0 $\rightarrow$ 1 change in layout image pixels, as the images are binary-valued.
    \end{enumerate}
    
    \begin{figure}[t]
        \centering
        \includegraphics[width=0.95\textwidth]{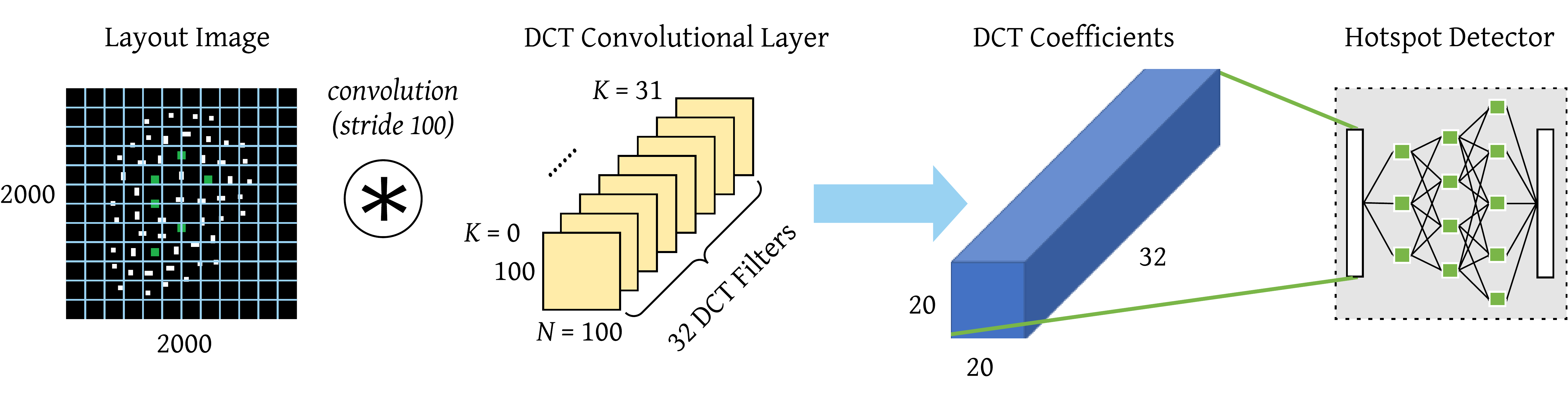}
        \caption{Illustration of end-to-end hotspot detection with DCT implemented as a convolutional layer.}
        \label{fig:dct}
    \end{figure}
    
    \subsubsection{DCT as a convolution layer in the network}
    Our solution to this problem is to implement the DCT computation as a convolution layer of the neural network such that the combined network works in an end-to-end fashion.
    The network takes in layout images as inputs; this allows us to perturb image pixels while also incorporating the attack constraints.
    This idea comes from Equation \ref{eqn:dct}, where we observed that the summation and element-wise product can be realized directly as a convolution layer of the CNN (without adding bias). % naturally be done by a convolution operation, which can
    The weights of a DCT filter for calculating the $K$-th DCT coefficient is obtained as shown in \autoref{equ:dctfilter}:
    \begin{equation}\label{equ:dctfilter}
        W_K = \cos{\bigg[\frac{\pi}{N_1}\bigg(n_1 + \frac{1}{2}\bigg)k_1\bigg]}\cos{\bigg[\frac{\pi}{N_2}\bigg(n_2 + \frac{1}{2}\bigg)k_2\bigg]}
    \end{equation}
    Here $k_1$ and $k_2$ are the horizontal and vertical coordinates of the $K$-th DCT coefficient, and $n_1$ and $n_2$ are the horizontal and vertical coordinates of each weight of the filter. $N_1$ and $N_2$ are the width and height of the filter.
    Since the DCT computation operates on 100~$\times$~100 sub-blocks of each image, the DCT convolution layer will have filters of size (100, 100) and strides of 100. We illustrate this end-to-end network that combines the DCT computation and hotspot detection in \autoref{fig:dct}.
    
    %\todoblock{Some sort of figure here to represent the DCT layer}
    
    \subsubsection{Attack Process}
    With this new end-to-end network, the attacker can now explore the gradients of the network in terms of the image and guide placement of SRAFs to positions that have the highest impact on the network output prediction (shifting from hotspot to non-hotspot).
    We define the loss of the attack as the \textit{distance} between the prediction probability of perturbed hotspot layout and \textit{ideal} non-hotspot layout (i.e., a layout with perfect prediction probability of [1,~0]).
    This is represented as \autoref{Equ:loss} where $P_{hotspot}(x)$ is the probability that a layout $x$ is classified as hotspot.%  \todoblock{Equation for loss function here?}
    
    \begin{equation}
        \label{Equ:loss}
        loss\_adv(x) = (1 - P_{hotspot}(x) - 1)^2 + P_{hotspot}(x)^2 = 2P_{hotspot}(x)^2
    \end{equation}
    
    The attacker aims to keep minimizing the loss as they choose and add perturbations (i.e, SRAFs) iteratively, until at some point the perturbed layout image is predicted as non-hotspot. 
    As an attack parameter the attacker can choose a maximum number of SRAFs to insert, $T$.
    The detailed algorithm of the white-box attack is shown in Algorithm \ref{alg:white-atk}.
    
\begin{figure}[t]
        \centering
        \includegraphics[width=\textwidth]{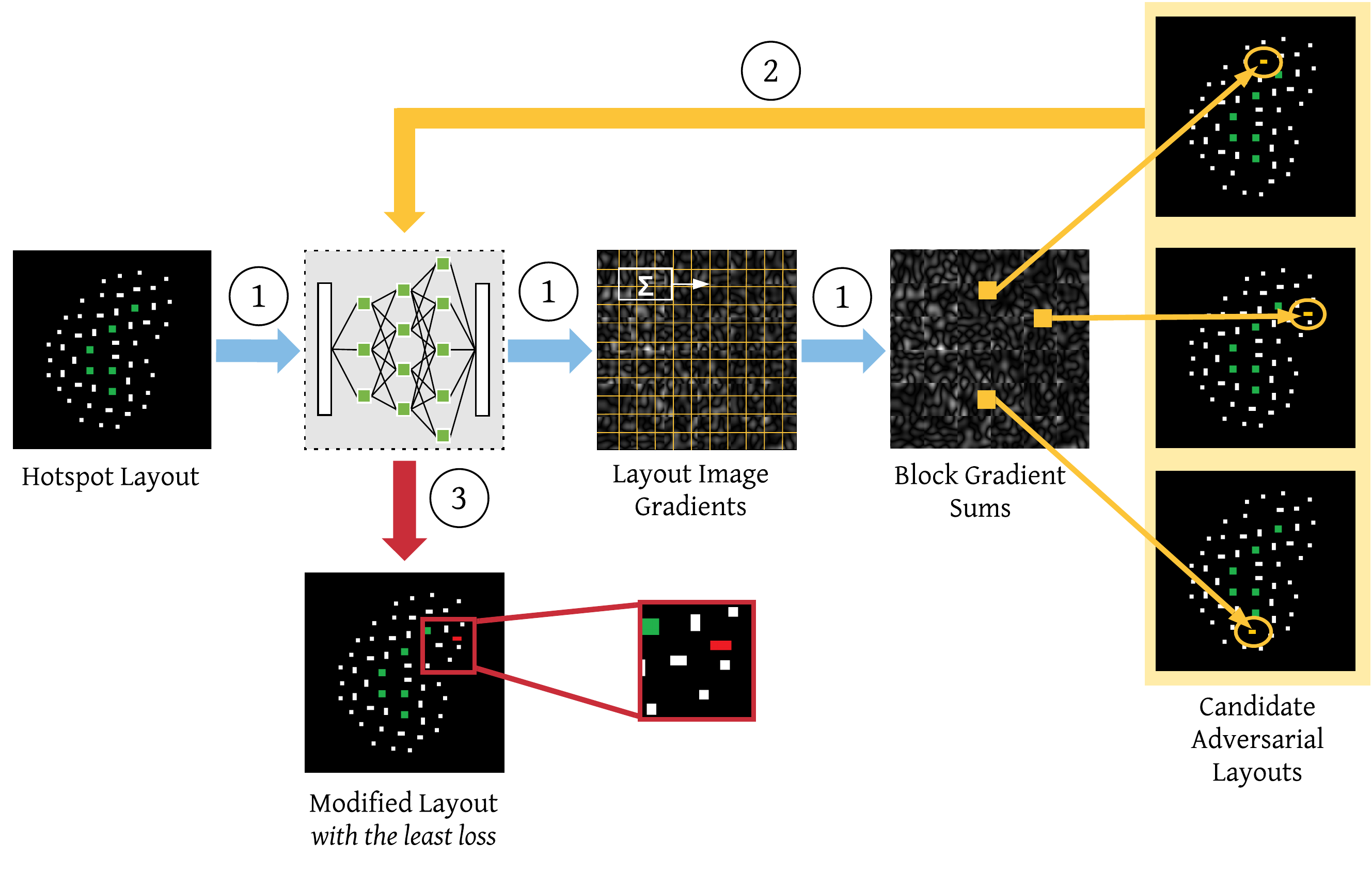}
        \caption{Illustration of the white-box attack process with one SRAF insertion.}
        \label{fig:whit_atk}
    \end{figure}

    To perform the attack, we first calculate the gradient of the loss function with respect to each pixel (line 4 in \autoref{alg:white-atk}).
    This gradient represents the amount of "influence" that a given pixel has on the final network prediction.
    However, since we are modifying \textit{blocks} of image pixels instead of a singe pixel (when we insert SRAFs), we sum the gradients of a potential perturbation block at each potential insertion location (line 5-10 in \autoref{alg:white-atk}). 
    %To the best of our knowledge, none of the existing adversarial perturbation attacks in literature have (or have needed to) 
    We illustrate this concept as \circled{1} in \autoref{fig:whit_atk}.
    This represents the "influence" that a perturbation has on the final network prediction when it is inserted in that location.
    Specifically, we are changing blocks of pixel values in the positive direction from 0 to 1, so the image block that has the largest negative sum of gradients will have the most influence in minimizing the loss.
    
    However, these gradient sums only reflect the influence for a \textit{small} change in the input. As we are shifting pixel values with a relatively large step (i.e., from 0 to 1) there is no guarantee that the largest negative sum of gradients will still have the most significant influence. Therefore, instead of picking the SRAF insertion that has the largest negative sum of gradients, we query the CNN for top-$n$ candidate perturbation blocks with the largest negative sum of gradients (line 11-14 in \autoref{alg:white-atk}).
    We refer to this $n$ as the attacker-specified \textit{check parameter}.
    Of these candidates, we pick the one that has the largest influence (i.e., that fools the network toward predicting a hotspot as non-hotspot) (line 15 in \autoref{alg:white-atk}).
    
    Our strategy for finding candidate SRAF insertion (\circled{2} in \autoref{fig:whit_atk}) is as follows.
    We define the location of SRAFs by the coordinate of its top-left corner point.
    We slide the SRAF over the center region of the layout (we leave a 200~nm boundary on each side of the layout image) with horizontal and vertical stride of 40~nm. This forms all the possible locations for potential perturbation addition.
    However, if any part of a location and its surrounding 40~nm has an existing pixel value of 1 (i.e., it is already occupied with an SRAF), or overlaps with any forbidden region, this location is marked as invalid for SRAFs. We set the loss of this location/shape pair to be $\infty$ (line 10 in \autoref{alg:white-atk}).
    In this way, we ensure that inserted SRAFs satisfy the attack constraints.
    
    \begin{algorithm}[H]
    \makeatletter
    \def\BState{\State\hskip-\ALG@thistlm}
    \makeatother
    \caption{White-box Attack}
    \label{alg:white-atk}
    \begin{small}
    \begin{algorithmic}[1]
    \State Input: original hotspot image $x$, white-box network function $F$ with DCT convolutional layer at the bottom, loss function $L$, image pixel indexing function $P$,  perturbation pattern set $S=height \times [width1, width2, \dots]$, surrounding spacing $d$, maximum number of perturbation addition $T$, check parameter $n$.
    %\Statex
    \State $t=0$
    % \BState \emph{Perturbation\ Insertion\ Loop}:
    \While{$t<T$}
    \State compute image gradient $x\_grad = \mathrm{d} L / \mathrm{d} x$
        \For{$shape$ in $S$}
            \For{each position $pos$ in $x$}
                \If {pixel values of $shape$ at $pos$ and its surrounding area $P(x, shape, pos, d)=0$}
                    \State $sum\_grad[shape, pos]=\sum P(x\_grad, shape, pos)$ \Comment{Sum gradients of the $shape$ area.}
                    % \State 
                \Else
                    \State $sum\_grad[shape, pos]=\infty$
                \EndIf
            \EndFor
        \EndFor
    
        \For{$i$ = 1 to $n$}
            \State Get $shape$ and $pos$ of the $i$-th smallest element of $sum\_grad$
            \State perturbed image $x' = x$ and set $P(x', shape, pos) = 1$
            \State compute loss $loss\_adv[i]=L(x')$ 
        \EndFor
        \State $shape,\ pos = \arg \min loss\_adv$ and set $P(x, shape, pos)=1$ \Comment{Insert perturbation.}
        \Statex
        \If {$F(x)$ is hotspot} 
            % \State \textbf{goto} \emph{Perturbation\ Insertion\ Loop}
            \State $t=t+1$
        \ElsIf {$F(x)$ is non-hotspot} 
            % \State print "Adversarial non-hotspot layout generated"
            \State \textbf{Return}: $x$ \Comment{Adversarial non-hotspot layout generated.}
            % \Else
            % \State print "Failed to generate non-hotspot layout"
        \EndIf
    % \Statex
    \EndWhile
    \State \textbf{Return}: $null$ \Comment{Otherwise, failed to generate non-hotspot layout within attacker-specified bound.}
    \end{algorithmic}
    \end{small}
    \end{algorithm}
    
    If the constraints are all satisfied, we consider this location to be valid.
    If the sum of the gradients at this location is one of the $n$ largest negative sums, we compute the loss for this layout image with the hypothetically inserted SRAF shape at this location (shown as \circled{3} of \autoref{fig:whit_atk}).
    Since the attacker has the flexibility to add six different shapes of SRAFs (width varies from 40, 50, 60, 70, 80 to 90~nm), they will iterate the gradient summation on all the possible locations for each of these perturbation shapes.
    In each iteration of adding one SRAF, one of the 6 shapes is added to the current layout such that it yields the lowest prediction probability for hotspot.
    
    The algorithm stops either when the network predicts the perturbed layout as non-hotspot (hotspot prediction probability $\leq$ 0.5), or when the number of inserted SRAFs has reached the maximum allowance and no adversarial non-hotspot layout is generated (line 16-20 in \autoref{alg:white-atk}). 
    
    % \textcolor{red}{Move this last sentence upwards in the text. Wherever possible, relate the text to specific lines in the algorithm, as in the examples above.}
    % got it, wilco
    % \todoblock{Need a figure of these steps, crossref each "step" in the prose with the accompanying subfigure}
    % They know the architecture, weights, etc. and wish to use this knowledge to either improve the likelihood of a successful misclassification, or speed-up their attack.

\subsection{Black-Box Attack}% -- a query based approach}
\begin{figure}[t]
        \centering
        \includegraphics[width=\textwidth]{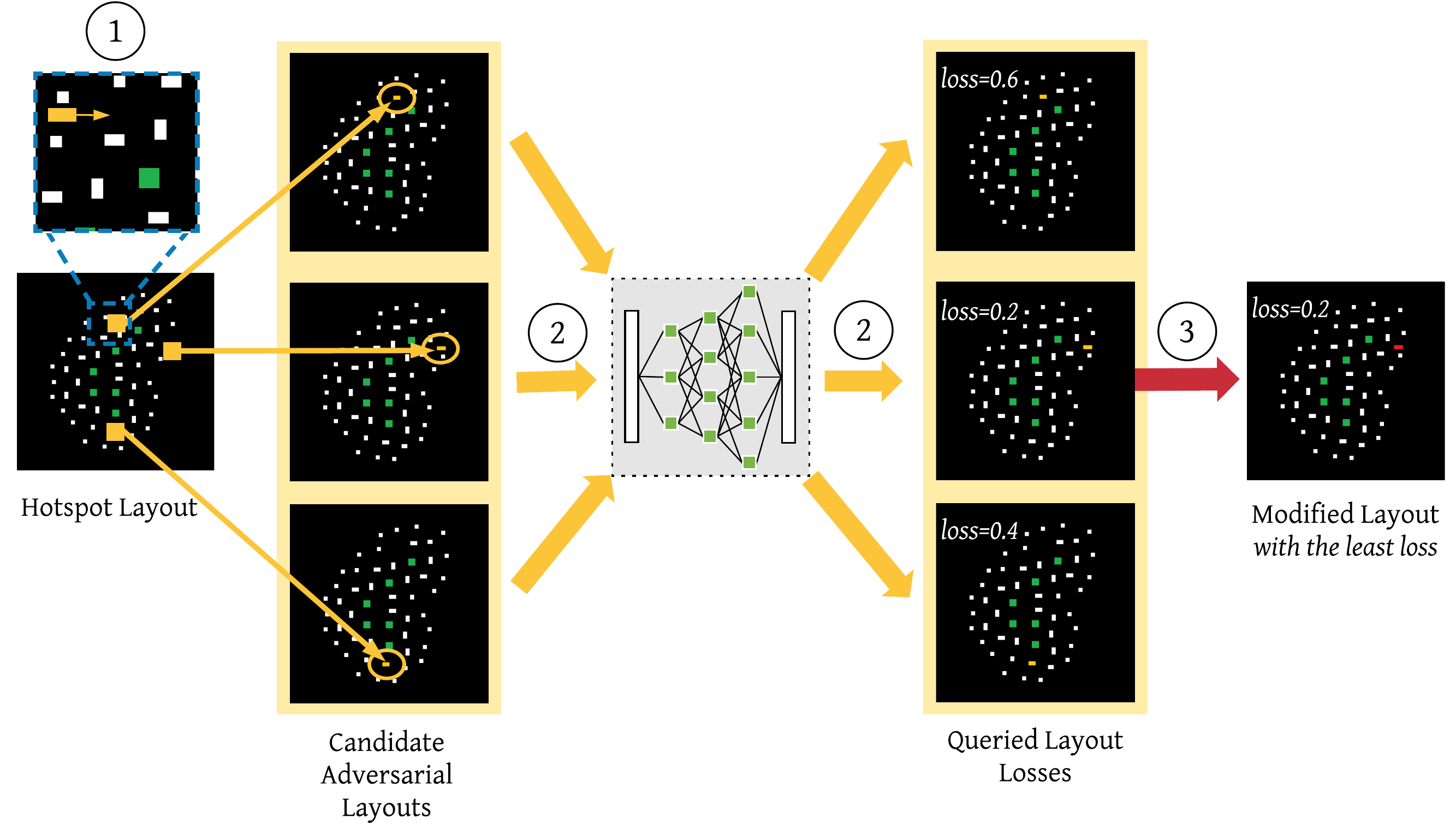}
        \caption{Illustration of the black-box attack process with one SRAF insertion.}
        \label{fig:black_atk}
    \end{figure}

    This attack explores the case where an attacker has less knowledge of the target network. With a black-box access to the network, the attacker can query the hotspot detector with computed DCT coefficients of a layout to obtain the output prediction probability. We illustrate the attack in \autoref{fig:black_atk}. Details of the black-box algorithm are shown in Algorithm \ref{alg:black-atk}.
    
    At a high-level, the black-box attack iteratively queries the detector with different SRAF shape and insertion location combinations. The attacker first adds a single SRAF.
    The attacker exhaustively examines all the possible valid locations for each valid SRAF shape (\circled{1} in \autoref{fig:black_atk}), and queries the network with DCT coefficients of the candidate modified layout (\circled{2} in \autoref{fig:black_atk}, line 4-10 in \autoref{alg:black-atk}).
    The location and perturbation that has the minimum loss is selected, using the same loss function as in \autoref{Equ:loss} (\circled{3} in \autoref{fig:black_atk}, line 11 in \autoref{alg:black-atk}).
    Further SRAFs are added in the same way. Like the white-box attack, the algorithm terminates either by returning a successful adversarial non-hotspot layout, or fails to produce an adversarial layout within the specified maximum number of inserted SRAFs.
    
    \begin{algorithm}[t]
    \makeatletter
    \def\BState{\State\hskip-\ALG@thistlm}
    \makeatother
    \caption{Black-box Attack}\label{alg:black-atk}
    \begin{small}
    \begin{algorithmic}[1]
    \State Input: original hotspot image $x$, DCT computation function $DCT$, black-box network function $F$, loss function $L$, image pixel indexing function $P$,  perturbation pattern set $S=height \times [width1, width2, \dots]$, surrounding spacing $d$, maximum number of perturbation additions $T$.
    \State $t=0$
    \While{$t<T$}
        \For{$shape$ in $S$}
            \For{each position $pos$ in $x$}
                \If {pixel values of $shape$ at $pos$ and its surrounding area $P(x, shape, pos, d)=0$}
                    \State perturbed image $x' = x$ and set $P(x', shape, pos) = 1$
                    \State compute loss $loss\_adv[shape, pos]=L(DCT(x'))$ 
                \Else
                    \State $loss\_adv[shape, pos]=\infty$
                \EndIf
            \EndFor
        \EndFor
        \State $shape, pos = \arg \min loss\_adv$ and set $P(x, shape, pos)=1$ \Comment{Insert perturbation.}
        \Statex
        \If {$F(DCT(x))$ is hotspot} 
            % \State \textbf{goto} \emph{Perturbation\ Insertion\ Loop}
            \State $t=t+1$
        \ElsIf {$F(DCT(x))$ is non-hotspot} 
            % \State print "Adversarial non-hotspot layout generated"
            \State \textbf{Return}: $x$ \Comment{Adversarial non-hotspot layout generated.}
            % \Else
            % \State print "Failed to generate non-hotspot layout"
        \EndIf
    \EndWhile
    \State \textbf{Return}: $null$ \Comment{Otherwise, failed to generate non-hotspot layout within attacker-specified bound.}
    \end{algorithmic}
    \end{small}
    \end{algorithm}
    
\section{Empirical Evaluation of Attack Success}
\label{sec:attack-results}
    \subsection{Experimental Setup}
    To investigate the implications of our proposals, as well as study the relationship between the attack efficacy and the target networks, we performed white-box and black-box attacks on both CNN-based hotspot detection networks.
    We run all experiments on a desktop computer with Intel CPU i9-7920X (12 cores, 2.90 GHz) and single Nvidia GeForce GTX 1080 Ti GPU.
    
    For the white-box attack, we conduct adversarial non-hotspot layout generation on 500 correctly classified hotspot layouts from the validation set (i.e., the layouts that were not used for training).
    As the black-box attack takes longer to perform, we generate adversarial non-hotspot layouts for 150 hotspot layouts.
    The attack success rate is the percentage of hotspot layouts that were perturbed such that they are misclassified as non-hotspot by the CNN. 
    Across all the experiments in this section, we consider a layout to be hotspot if the network prediction probability for hotspot is $\geq$ 0.5. 
    The average attack time is the end-to-end time (including querying the hotspot detector). 
    %this was recorded using Python built-in timer.
    We limit the maximum number of adversarial SRAF additions to 20, and the check parameter in the white-box attack is 180\footnote{this corresponds to \textasciitilde11\% of the total number of possible SRAF insertion candidates on average.}.
    We illustrate a selection of attack outputs and their corresponding verification results in \autoref{fig:adv_litho_example1} and \autoref{fig:adv_litho_example3}. 
    We present a summary of the results in \autoref{tab:results}.

    % \begin{table}
    % \centering
    % \caption{Summary of Attack Results for White-box and Black-box algorithms. For both attacks, the maximum number of SRAF insertions allowed ($T$) is 20. For the white-box attack, the check parameter ($n$) is 30. $T$ and $n$ are attacker-defined parameters, as explained in Algorithms \ref{alg:white-atk} and \ref{alg:black-atk}.}
    % \label{tab:results}
    % \begin{tabular}{@{}lrrrrrr@{}}
    % \toprule
    % Attack & \multicolumn{3}{c}{White-box} & \multicolumn{3}{c}{Black-box}  \\ \cmidrule(l){2-4} \cmidrule(l){5-7}
    % Network & \multicolumn{1}{c}{A} & \multicolumn{1}{c}{B} & \multicolumn{1}{c}{C} & \multicolumn{1}{c}{A} & \multicolumn{1}{c}{B} & \multicolumn{1}{c}{C} \\ \midrule
    % Attack success rate & 99.7\% & 85.5\% & 33.4\% & 99.7\% & 93.3\% & 63.3\% \\
    % Average attack time per layout & 8.6 s & 45.1 s & 71.7 s & 350.5 s & 677.3 s & 1031.6 s \\
    % % Maximum num. of SRAFs allowed ($T$) & 20 & 20 & 20 & 20 & 20 & 20 \\
    % % Check parameter ($n$) & 30 & 30 & 30 & - & - & - \\
    % Average num. of SRAFs added & 5.3 & 8.3 & 9.8 & 4.1 & 7.3 & 8.5 \\
    % Average area of SRAFs added & 0.3\% & 0.5\% & 0.5\% & 0.3\% & 0.5\% & 0.5\% \\ \bottomrule
    % \end{tabular}
    % \end{table}
    
    \begin{table}[t]
    \centering
    \caption{Summary of attack results for white-box and black-box algorithms. For both attacks, the maximum number of SRAF insertions allowed ($T$) is 20. For the white-box attack, the check parameter ($n$) is 180. $T$ and $n$ are attacker-specified parameters, as explained in Algorithms \ref{alg:white-atk} and \ref{alg:black-atk}.}
    \label{tab:results}
    \begin{tabular}{lrrrr}
    \hline
    Attack & \multicolumn{2}{c}{White-box} & \multicolumn{2}{c}{Black-box} \\ \cmidrule(l){2-3} \cmidrule(l){4-5}
    Network & \multicolumn{1}{c}{A} & \multicolumn{1}{c}{B} & \multicolumn{1}{c}{A} & \multicolumn{1}{c}{B} \\ \hline
    Attack success rate & 99.7\% & 85.5\% & 99.7\% & 93.3\% \\
    Average attack time per layout & 8.6 s & 45.1 s & 350.5 s & 677.3 s \\
    Average number of SRAFs added & 5.3 & 8.3 & 4.1 & 7.3 \\
    Average area of SRAFs added & 0.3\% & 0.5\% & 0.3\% & 0.5\% \\ \hline
    \end{tabular}
    \end{table}
    
    \subsection{Attack Results}
    %As can be seen for the white-box attack, 
    The most successful attack was on Network A, where we achieved a 99.7\% attack success (498 hotspot layouts were made to be classified as non-hotspot by the CNN).
    %We observe that 
    The white-box attack success rate drops between our attack of Network A and B (a decrease of 14.2\%).
    One explanation is that the complex Network B has learned more about the characteristics of hotspots, and therefore is more challenging to fool; this is consistent with prior findings of neural network robustness \cite{madry_towards_2017}.
  
    These trends can be observed in the average time taken to generate an adversarial layout in the white-box attack.
    Network B required on average \textasciitilde6$\times$ more time than the white-box attack on Network A.
    The extra time for the white-box attack on Network B to produce a successful perturbed layout is partially due to the increased feedforward computation on more layers (higher overhead in query-time during the attack).
    Similarly, the average number of SRAFs inserted is greater for the more complex Network B compared to the simpler Network A.
    % We infer that in the biased case, the decision boundary 
    
\begin{figure}[t]
        \centering
        \subfigure[White-box Attack on Network A]{\label{fig:hist_a_w}\includegraphics[width=0.45\textwidth]{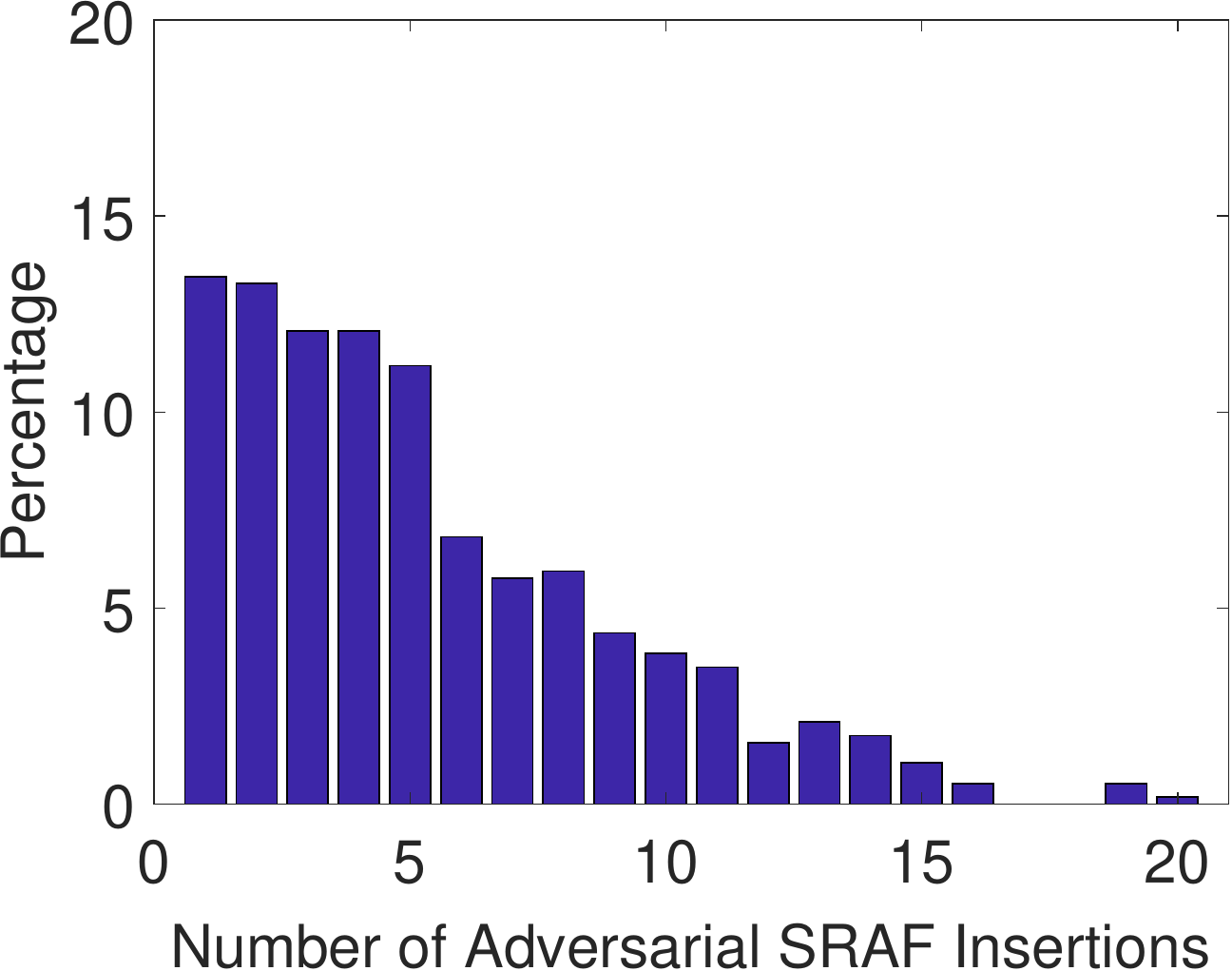}}
         \hspace{1.5em}
        \subfigure[Black-box Attack on Network A]{\label{fig:hist_a_b}\includegraphics[width=0.45\textwidth]{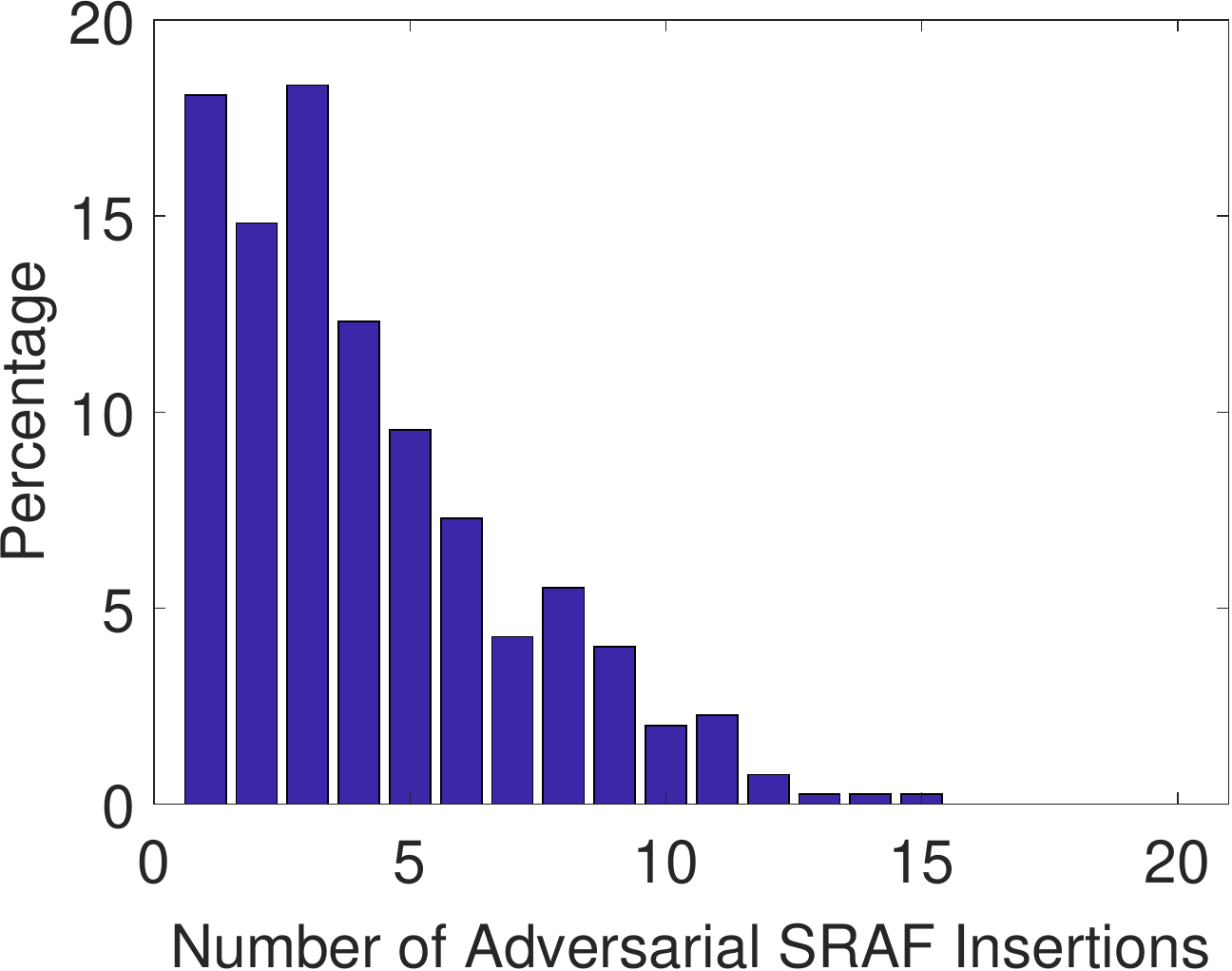}}
        \subfigure[White-box Attack on Network B]{\label{fig:hist_b_w}\includegraphics[width=0.45\textwidth]{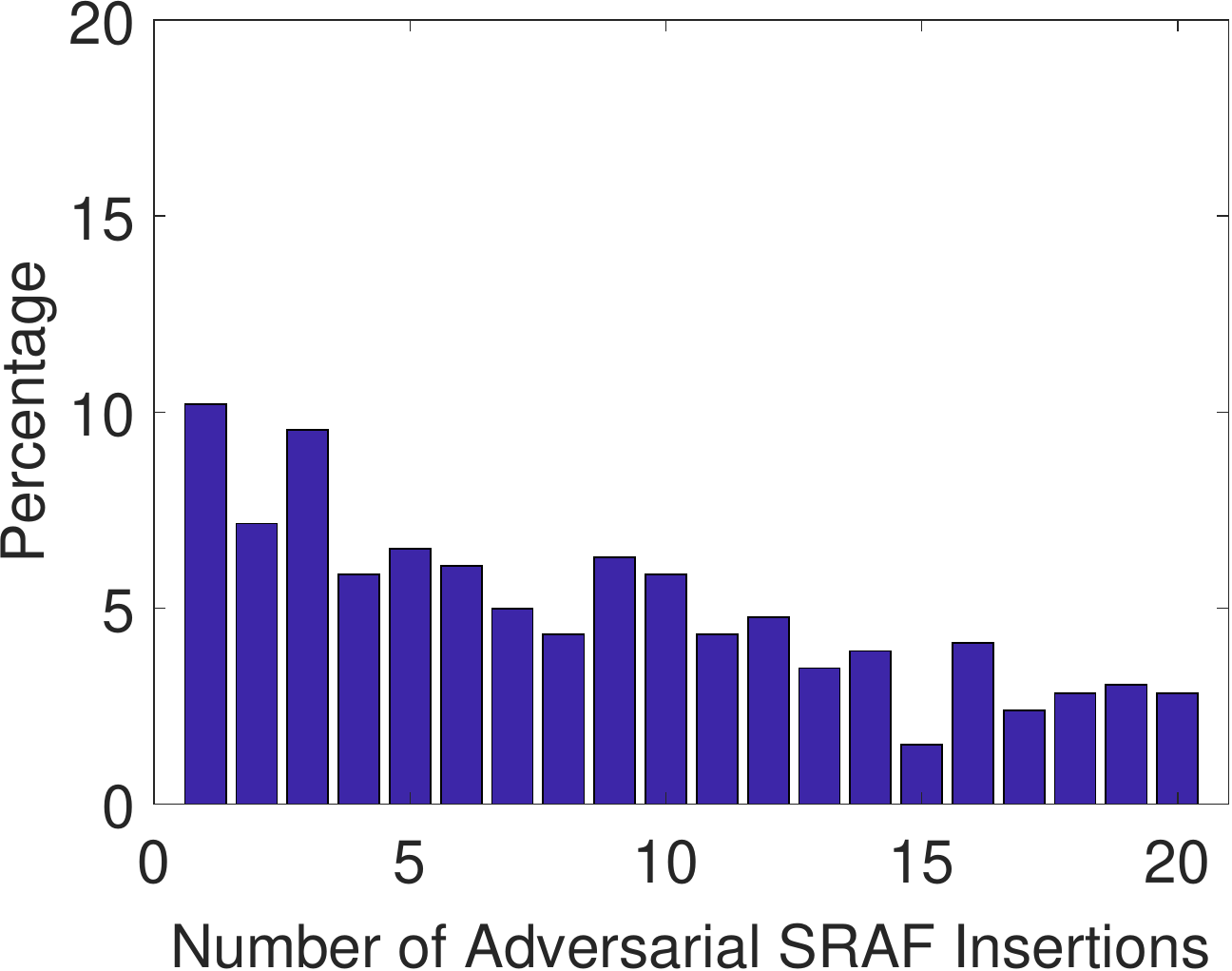}}
        \hspace{1.5em}
        \subfigure[Black-box Attack on Network B]{\label{fig:hist_b_b}\includegraphics[width=0.45\textwidth]{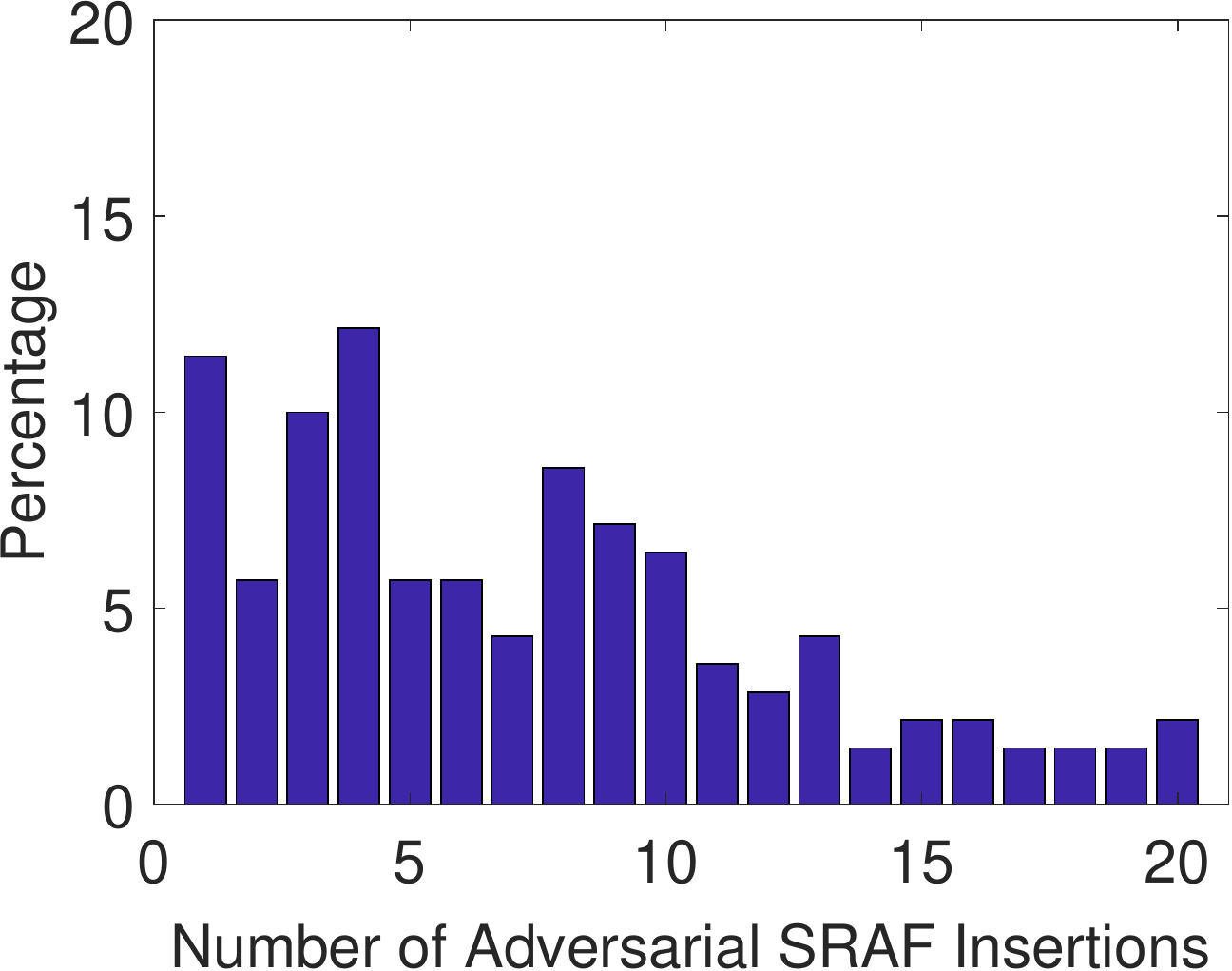}}
        % \subfigure[White-box Attack on Network C]{\label{fig:hist_c_w}\includegraphics[width=0.45\textwidth]{Fig/Experiment/C_cnt_w.pdf}}
        % \hspace{1.5em}
        % \subfigure[Black-box Attack on Network C]{\label{fig:hist_c_b}\includegraphics[width=0.45\textwidth]{Fig/Experiment/C_cnt_b.pdf}}
        \caption{Histograms of percentages of different number of adversarial SRAF insertions by white-box attack on Network A (a) and B (c). Histograms of percentages of different number of adversarial SRAF insertions by black-box attack on Network A (b) and B (d).}% and C (f).}
        \label{fig:hist_adv}
    \end{figure}

\begin{figure}[t]
    \centering
    \subfigure[Original Layout]{\label{fig:orig_layout_1}\includegraphics[width=0.28\textwidth]{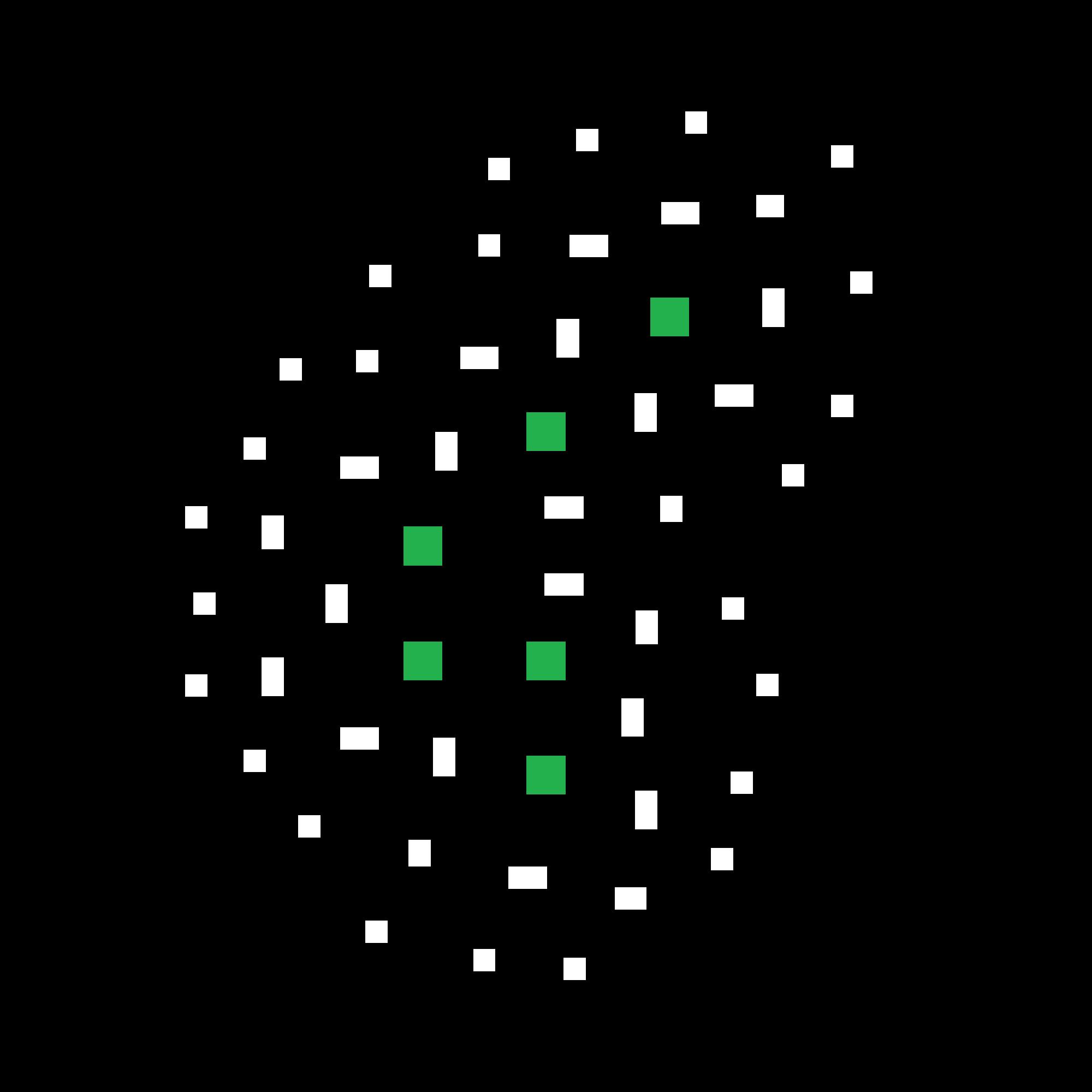}}  \hspace{1em}
    \subfigure[Adversarial Layout (white-box)]{\label{fig:w_layout_1}\includegraphics[width=0.28\textwidth]{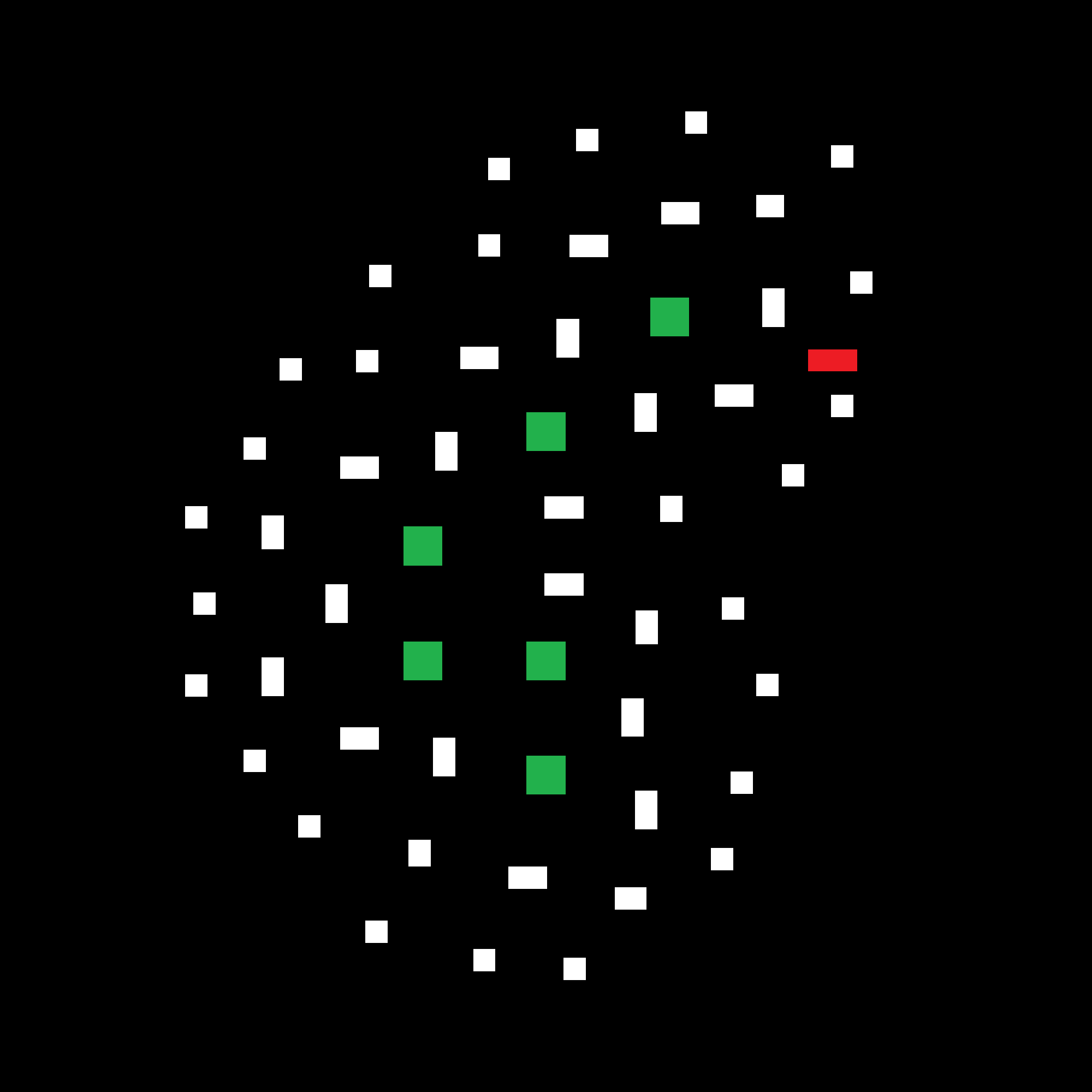}} \hspace{1em}
    \subfigure[Adversarial Layout (black-box)]{\label{fig:b_layout_1}\includegraphics[width=0.28\textwidth]{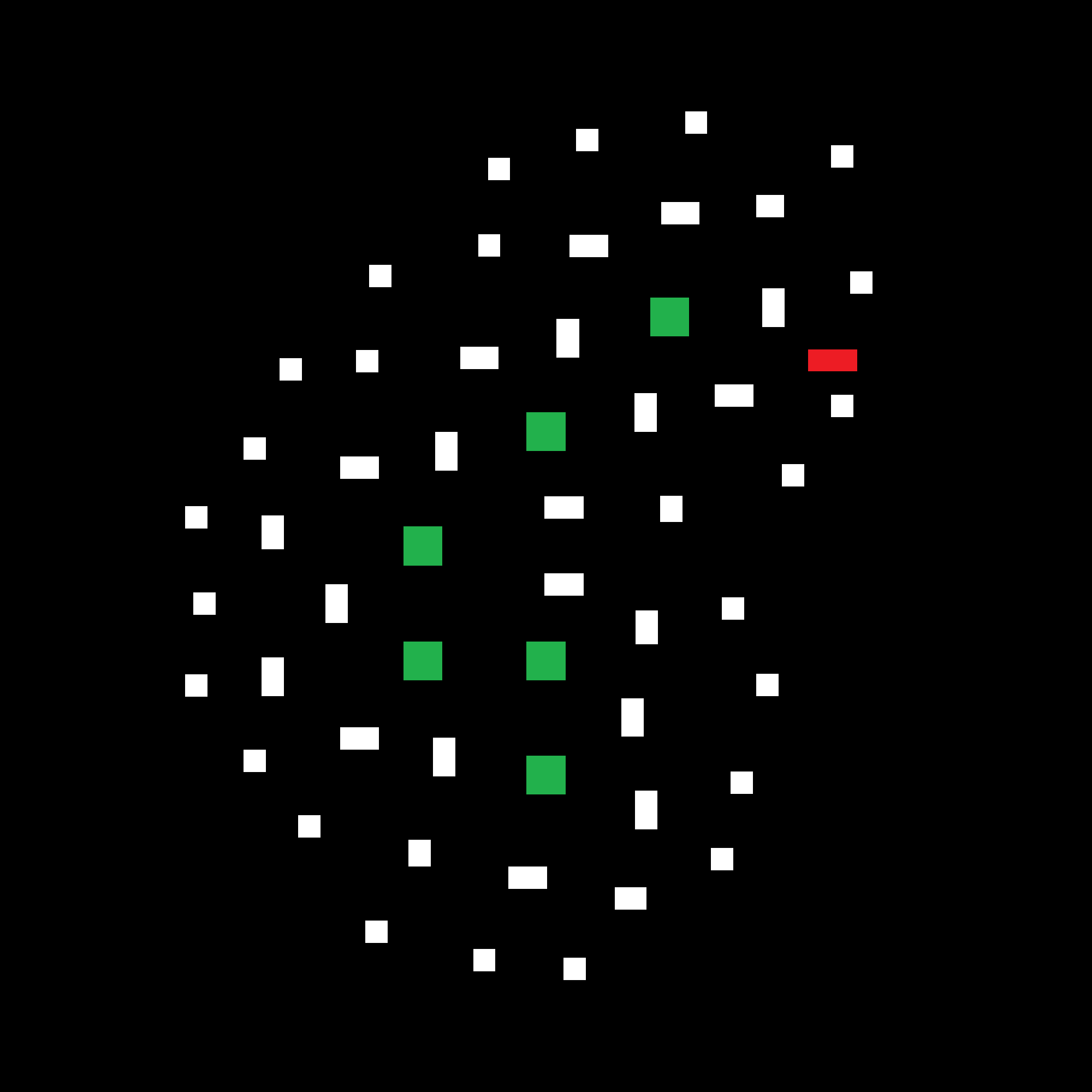}}
    \subfigure[Lithography Simulation of \ref{fig:orig_layout_1}]{\label{fig:orig_litho_1}\includegraphics[width=0.28\textwidth]{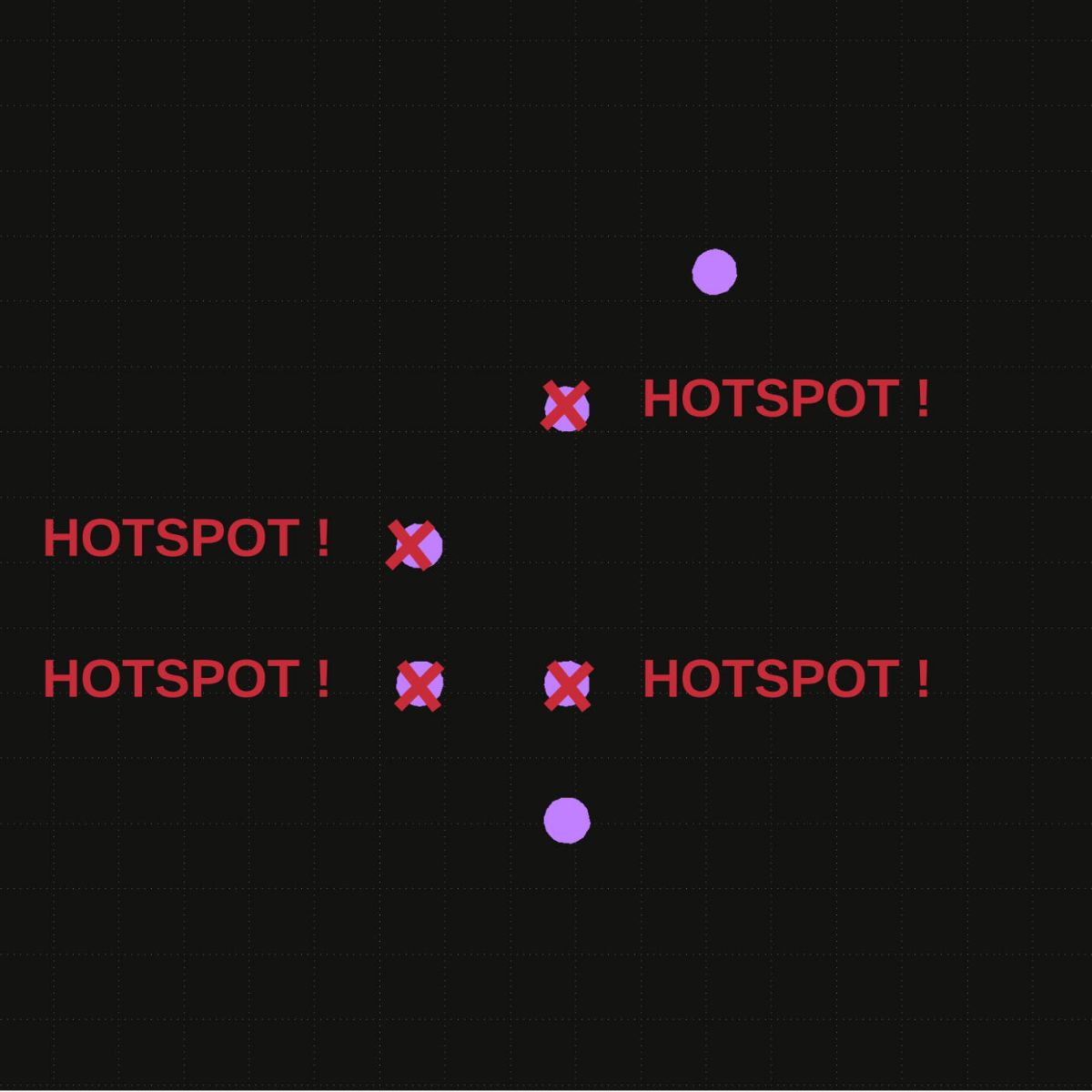}} \hspace{1em}
    \subfigure[Lithography Simulation of \ref{fig:w_layout_1}]{\label{fig:w_litho_1}\includegraphics[width=0.28\textwidth]{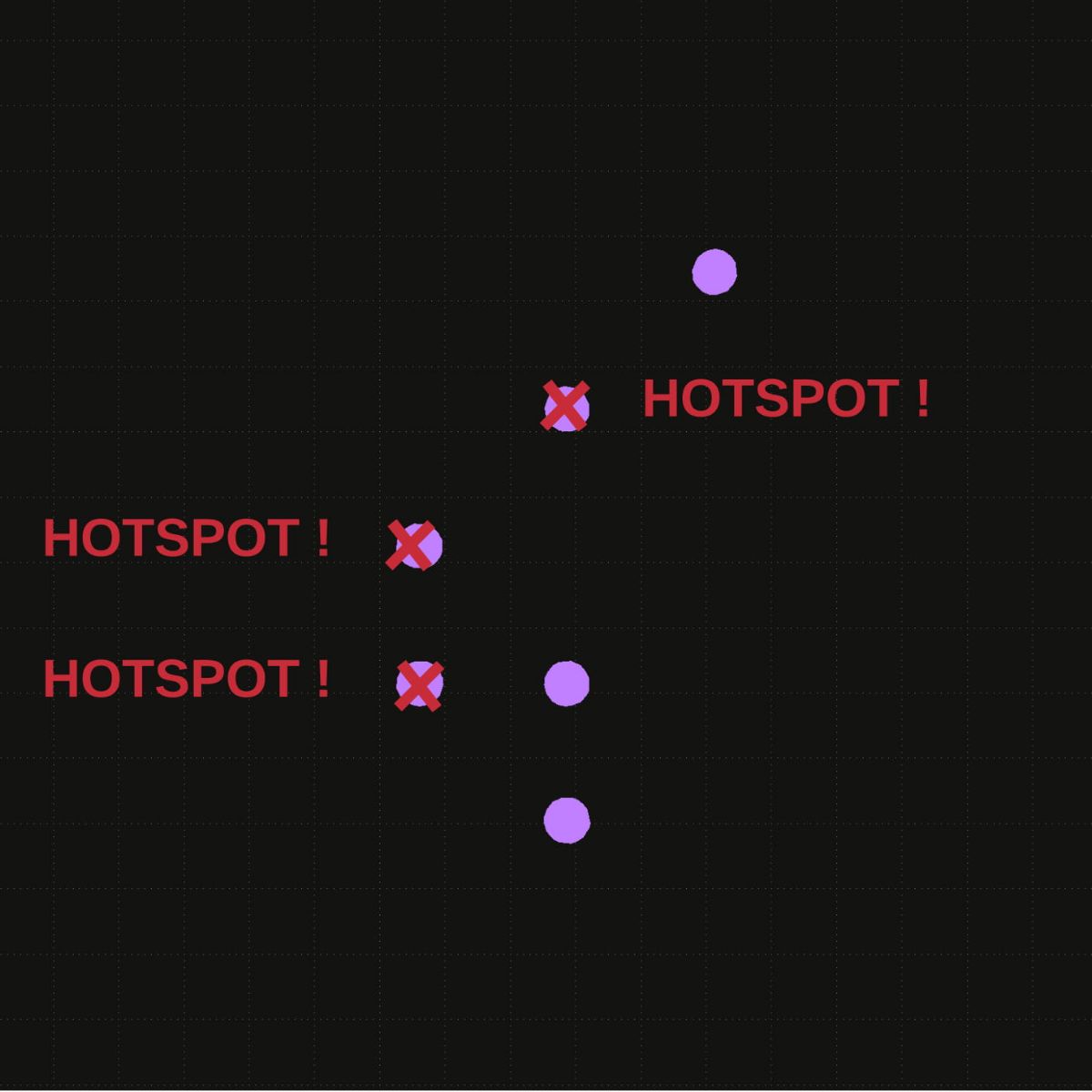}} \hspace{1em}
    \subfigure[Lithography Simulation of \ref{fig:b_layout_1}]{\label{fig:b_litho_1}\includegraphics[width=0.28\textwidth]{Fig/Experiment/epe_10299_white.jpg}}
    \caption{%
        White-box and black-box attack outputs -- Example 1. These examples feature a single inserted SRAF. In the layout images (a-c), the vias are colored green, the original SRAFs are white, and the adversarial SRAFs are red. In the lithography simulation outputs, the vias are shown in purple and hotspots are marked with a cross and labelled with "HOTSPOT!".
        % Attack Output Example 1. In Layouts: vias are green, original SRAFs in white, and adversarial SRAFs in red. In lithoography simulations (lith-sim): vias are purple, and hotspots are marked with a cross and labelled with "HOTSPOT!".
        }
    \label{fig:adv_litho_example1}
    \end{figure}

    \autoref{fig:hist_a_w} and \autoref{fig:hist_b_w} show the percentage of layouts that were successfully perturbed by a given number of SRAF insertions for the white-box attack.    
    In all cases, the minimum number of SRAFs that needed to be added to cause misclassification was 1, and an example of this is shown in \autoref{fig:adv_litho_example1}.
    Of the layouts that were successfully perturbed to appear as non-hotspot in each attack, \textasciitilde13\% required only one inserted SRAF to fool Network A, and \textasciitilde10\% for Network B. %, and \textasciitilde5\% for Network C.
    Furthermore, 50\% of the perturbed layouts could fool Network A with 4 or fewer inserted SRAFs.
    For Network B, 50\% of the perturbed layouts had 7 or fewer inserted SRAFs.%, while for Network C, 50\% had 9 or fewer inserted SRAFs.
    
    % \subsection{Black-Box Attack}
    Looking to the black-box attack results, we notice the same general trends as the white-box attack, where the simpler Network A is attacked most successfully, while the complex Network B exhibits greater resilience.
    \autoref{fig:hist_a_b} and \autoref{fig:hist_b_b} show the percentage of layouts that were successfully perturbed by a given number of SRAF insertions for the black-box attack. As with the white-box attack, the black-box attack yielded adversarial layouts with as few as one inserted SRAF (an example is shown in \autoref{fig:adv_litho_example1}). Of the layouts that were successfully perturbed to appear as non-hotspot, \textasciitilde18\% required one inserted SRAF to fool Network A and \textasciitilde11\% required one SRAF to fool Network B.  %, and \textasciitilde7\% required one SRAF to fool Network C.
    In the black-box attack on Network A, 50\% of the adversarial layouts required 3 or fewer added SRAFs.
    For Network B, 6 or fewer SRAFs were required in 50\% of the adversarial layouts.%, while for Network C, 50\% of the adversarial layouts required 8 or fewer.
    
    \subsection{Comparison and Observations}
    
    % \todoblock{Talk about white vs black, different number of SRAFs, "natural" resistance. Talk about how the whitebox thing can be useful from a defenders perspective?? Then crossref to defense section.}
    
\begin{figure}[t]
    \centering
    \subfigure[Original Layout]{\label{fig:orig_layout_3}\includegraphics[width=0.28\textwidth]{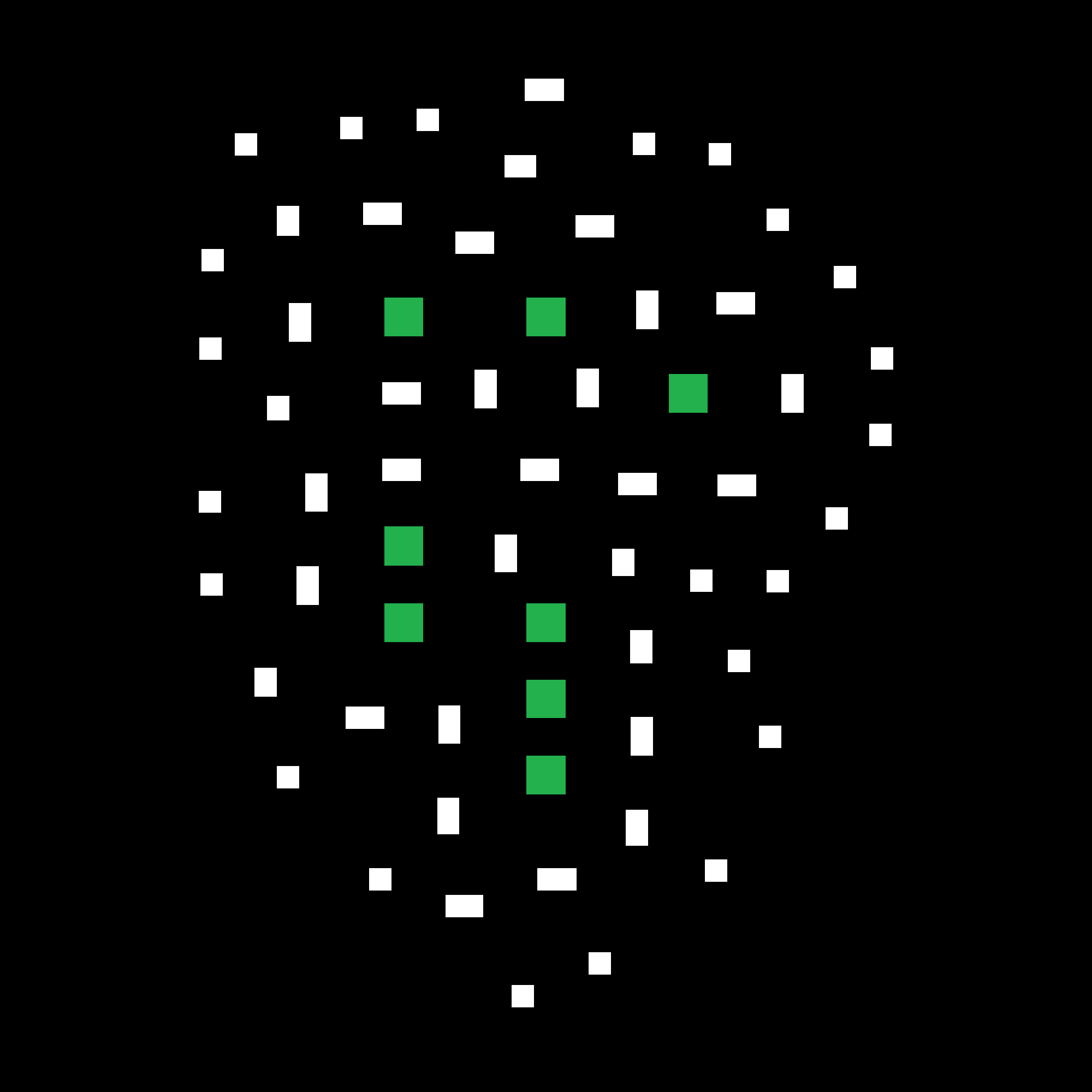}} \hspace{1em}
    \subfigure[Adversarial Layout (white-box)]{\label{fig:w_layout_3}\includegraphics[width=0.28\textwidth]{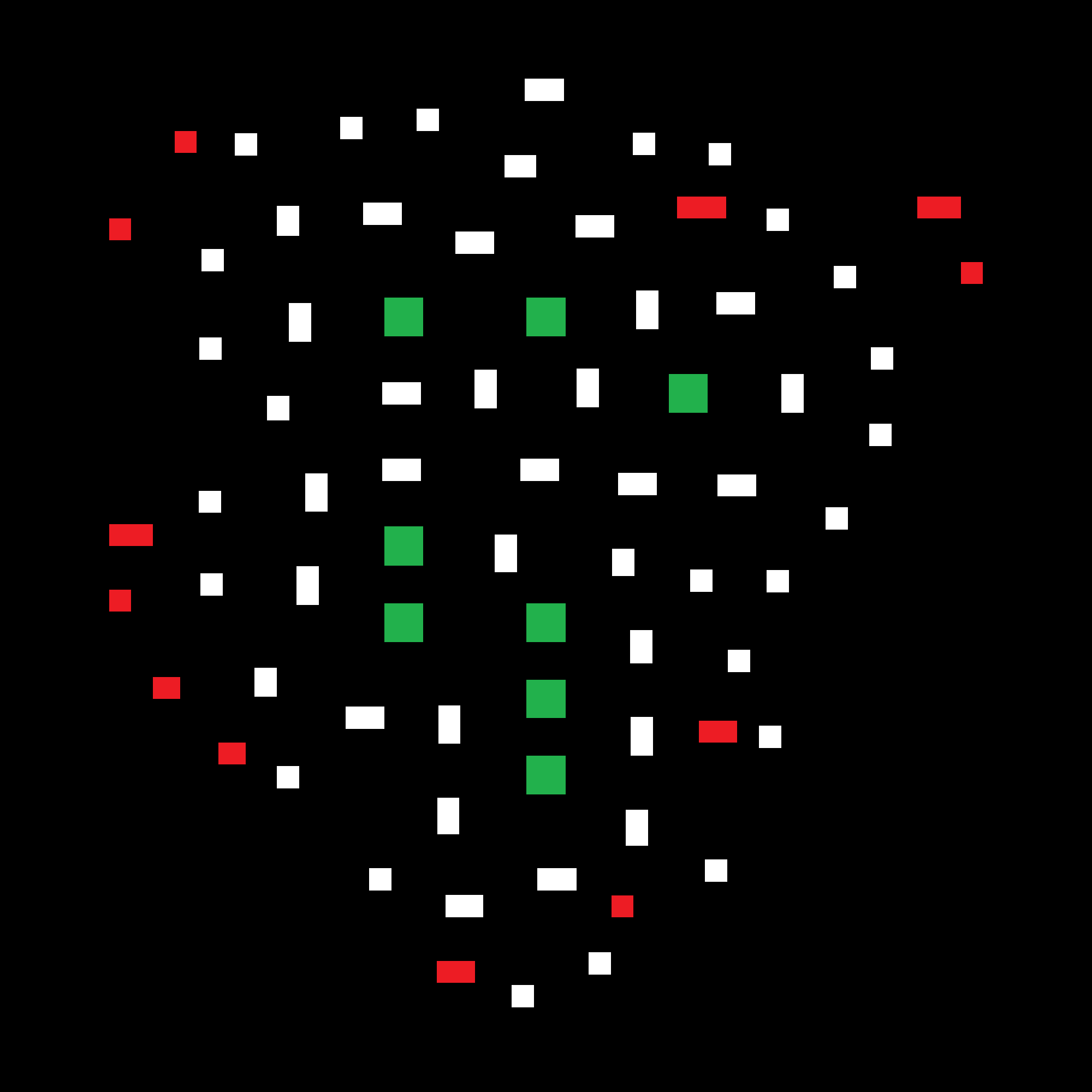}} \hspace{1em}
    \subfigure[Adversarial Layout (black-box)]{\label{fig:b_layout_3}\includegraphics[width=0.28\textwidth]{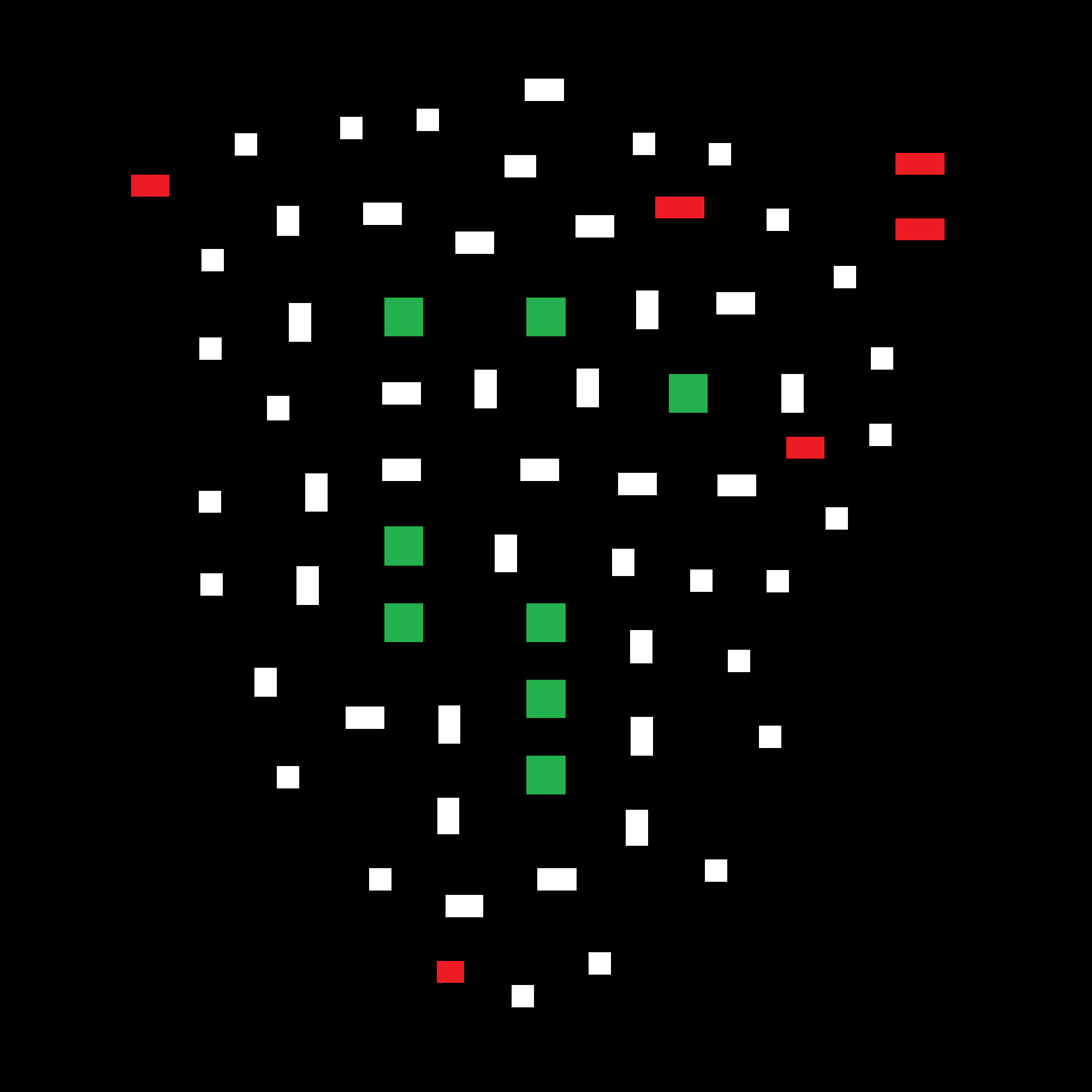}}
    \subfigure[Lithography Simulation of \ref{fig:orig_layout_3}]{\label{fig:orig_litho_3}\includegraphics[width=0.28\textwidth]{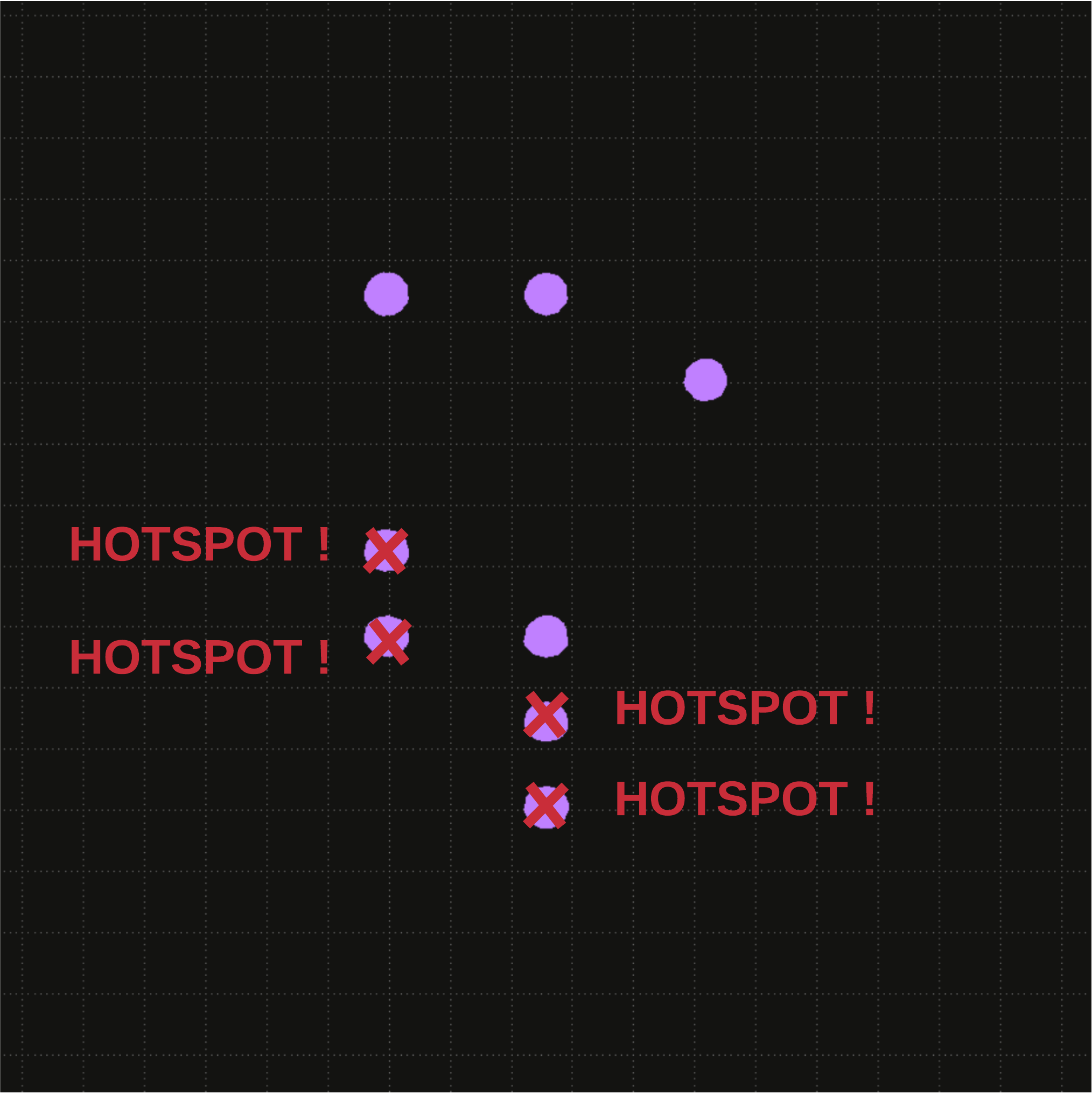}} \hspace{1em}
    \subfigure[Lithography Simulation of \ref{fig:w_layout_3}]{\label{fig:w_litho_3}\includegraphics[width=0.28\textwidth]{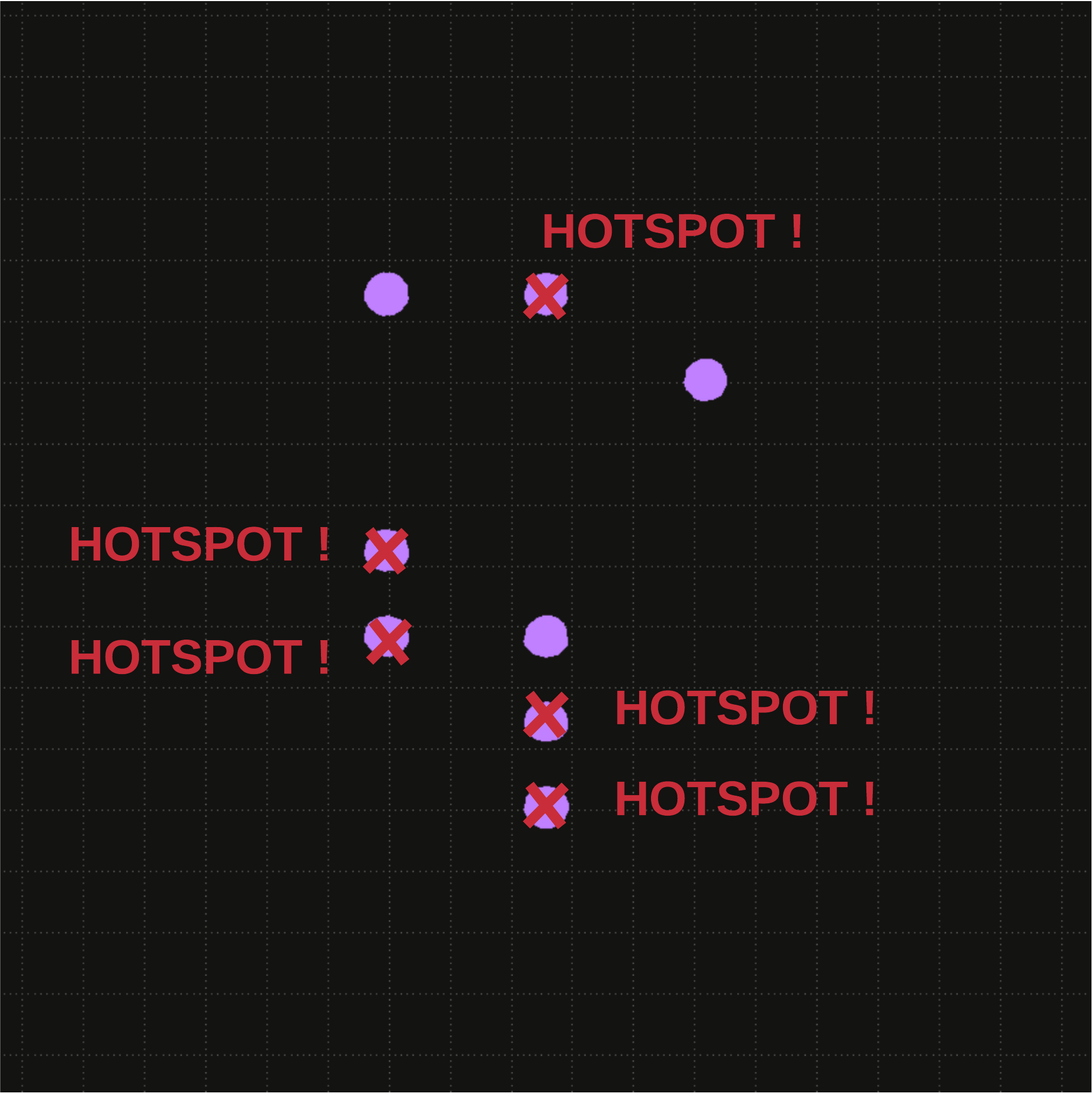}} \hspace{1em}
    \subfigure[Lithography Simulation of \ref{fig:b_layout_3}]{\label{fig:b_litho_3}\includegraphics[width=0.28\textwidth]{Fig/Experiment/epe_2945_orig.pdf}}
    \caption{%
        White-box and black-box attack outputs -- Example 2. These examples feature multiple inserted SRAFs. In the layout images (a-c), the vias are colored green, the original SRAFs are white, and the adversarial SRAFs are red. In the lithography simulation outputs, the vias are shown in purple and hotspots are marked with a cross and labelled with "HOTSPOT!".
        %The same color scheme from \autoref{fig:adv_litho_example1} is reused.
        %In the layout images (a-c), the vias are colored green, the original SRAFs are white, and the adversarial SRAFs are red. In the lithography simulation outputs (Lithography Simulation), the vias are shown in purple and hotspots are marked with a cross and labelled with "HOTSPOT!".
        }
    \label{fig:adv_litho_example3}
    \end{figure}

    There are a number of commonalities and differences between the results of white-box and black-box attacks.
    Both attacks produced adversarial layouts with only one added SRAFs for a number of layouts. 
    Attacks on the simpler Network A had a higher attack success rate in both white-box and black-box cases.
    There is also a notable increase in the average number of adversarial SRAFs added; the white-box attack requires 1-2 more SRAFs on average.
    An example of different SRAF insertions produced by the white-box and black-box attack can be seen in \autoref{fig:adv_litho_example3}.
    
    Of particular interest to an attacker is the feasibility of the attack in terms of computational overhead.
    One measure of this is the time taken to generate an adversarial layout.
    Our results show that the more successful black-box attack is 10$\times$ slower than the white-box attack.
    The trade-off for such high attack success is increased time and computation requirements.
    This can be explained by the number of times the attacker needs to query the CNN-based detector.
    In the white-box attack, the attacker only needs to query the network $n+1$ times in each iteration. The first query is incurred when using the network to compute the loss function gradients for each pixel. The subsequent $n$ queries obtain the prediction probabilities for candidate adversarial layouts. 
    When $n$ is set to the maximum number of possible shape/position combinations, white-box is equivalent to the black-box attack.
   
    As an attacker, one could tune the \textit{check parameter}, $n$, to balance the attack success rate against the computation resources required.
    Thanks to the gradient information used to guide the placement of adversarial SRAFs in the white-box algorithm, $n$ need not be too large to achieve reasonable or even comparable attack success rate as the black-box attack, while taking advantage of up to a 10$\times$ attack time reduction.
    This can also be useful for defenders, as we discuss in \autoref{sec:defense}.
    
    \subsection{Do Adversarial Perturbations Fix Hotspots?}% and Observations}
        Given the success of our adversarial perturbations, a natural question to ask is whether, using perturbations, we are actually fixing hotspots instead of misleading the designer. 
        To answer this question, we performed lithography simulation of the adversarial layouts to confirm the hypothesis that inserting only a few SRAFs does not drastically improve/fix hotspots but instead cause misclassification in most cases.
        We used the same experimental settings as those for ascertaining the ground truth labels of the original dataset (described in \autoref{sec:case-study}). 
        Examples of original and adversarial layouts, as well their simulation outputs, are shown in \autoref{fig:adv_litho_example1} and \autoref{fig:adv_litho_example3}.
        The simulations revealed that in the majority of cases, our adversarial layouts still produced layout defects.
        In the white-box attack on Network A, 84.4\% of the adversarial layouts that the network classified as non-hotspot were verified as hotspot, while in the same attack on Network B, 77.4\% of the adversarial layouts that were classified as non-hotspot were verified as hotspot.
        In the black-box attack, the verification rate was similar, where 86.7\% and 77.9\% of the layouts that fooled Network A and B (respectively) were verified as hotspot.
        
        When we examined the lithography simulation outputs we found instances where the number of hotspots in a layout increased, decreased, and stayed the same.
        An example of an instance where the inserted SRAFs added hotspots is shown in \autoref{fig:adv_litho_example3}(e), instances where the inserted SRAFs led to less hotspots is shown in \autoref{fig:adv_litho_example1}(e)(f), and an instance where the inserted SRAFs did not change the hotspot number is shown in \autoref{fig:adv_litho_example3}(f).

\section{Towards a More Robust Network}
\label{sec:defense}
    \subsection{Iterative Adversarial Retraining}
    So far we have shown that the white-box and black-box attacks are effective. Since this implies a feasible threat to deep learning based CAD, there is a need to investigate and propose countermeasures. As such, we propose a strategy to increase the robustness of CNN-based hotspot detectors.
    The main aim is to reduce the attack success rate without compromising hotspot detection accuracy. The approach can be integrated into the initial training process for the CNN-based network and is a type of adversarial retraining, as proposed in \cite{tramer_ensemble_2017}.
    
    First, let us assume that the defender knows the risks of adversarial perturbations on hotspot detectors.
    Intuitively they can make the trained network robust by including adversarial layouts into the training dataset but with true hotspot labels, and then retrain their detector using the usual methods \cite{tramer_ensemble_2017}.
    To diversify the training data set, the defender can adopt the attacker's methodology to proactively generate their own adversarial layouts and include them after verifying the true labels using lithography simulation.
    For robustness, the defender can repeat the adversarial retraining to suppress the success rate of adversarial attacks on the robust retrained network.% until a pre-defined threshold is achieved.
    
\begin{figure}[t]
        \centering
        \includegraphics[width=0.9\textwidth,trim=20 15 20 15,clip]{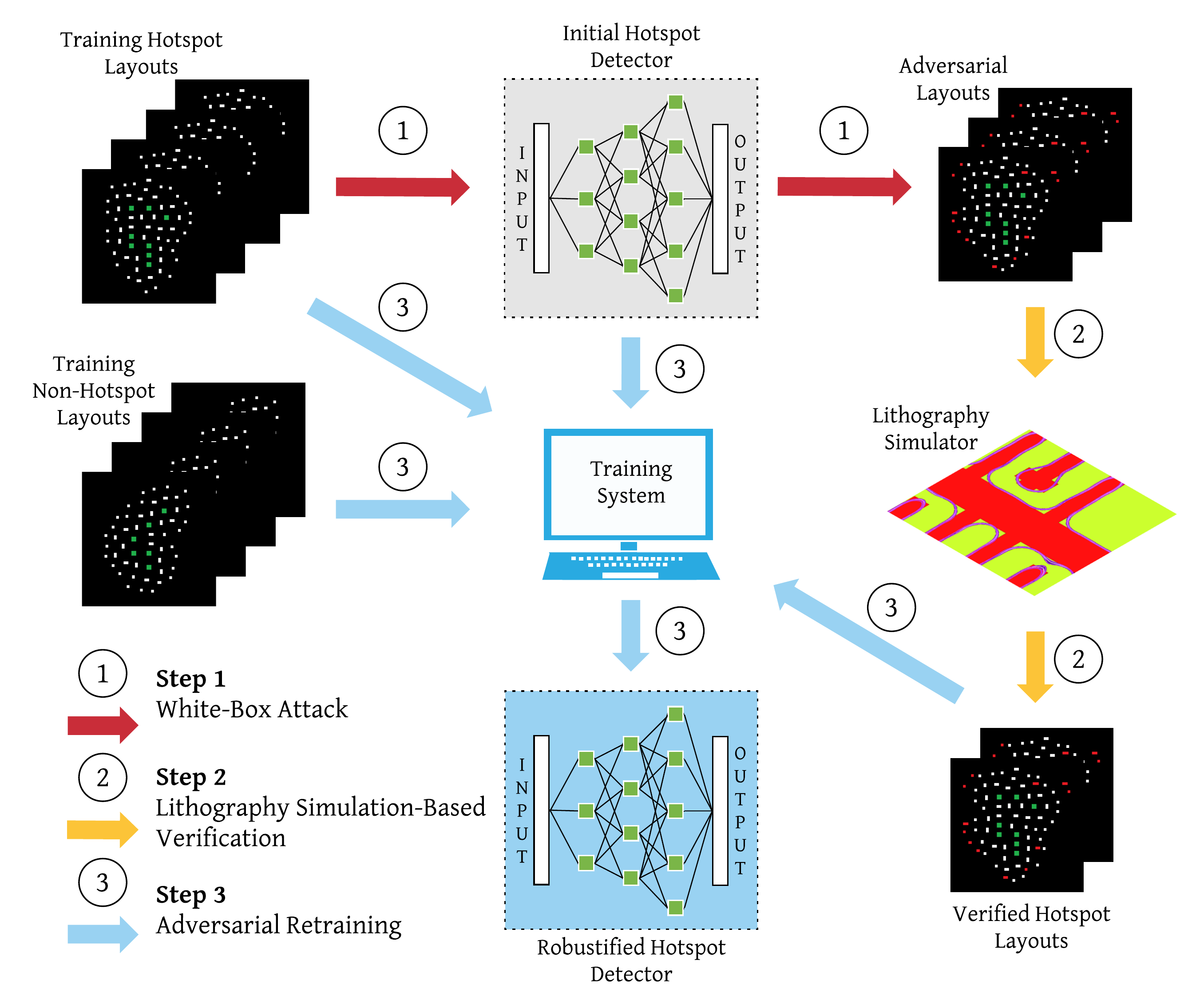}
        \caption{Overview of the adversarial retraining process.}
        \label{fig:adv_retrain}
    \end{figure}

    In practice, as we showed in \autoref{sec:attack-results}, the black-box attack achieves the highest possible attack success rate. Hence, it would make sense to make the network robust using adversarial layouts produced by this attack.
    However, given that it can be 10$\times$ slower than the white-box attack, this is less feasible under time and computation resource constraints.
    Therefore, while the white-box attack may have lower success rate in some occasions, it is more efficient in generating adversarial layouts.
    We adopt the white-box attack as part of the defender's strategy, and this provides ample training data and robustification results.
    Adversarial retraining is shown in Algorithm \ref{alg:adv-retrain} and the flow is shown in \autoref{fig:adv_retrain}.
    
    %\begin{comment}        
    \begin{algorithm}[t]
    \makeatletter
    \def\BState{\State\hskip-\ALG@thistlm}
    \makeatother
    \caption{Adversarial Retraining}\label{alg:adv-retrain}
    \begin{small}
    \begin{algorithmic}[1]
    \State Input: Training data $D_{train}$, Training hotspot data $D_{train, hotspot}$, network function $F$, adversarial non-hotspot layout generation function $Attack$, lithography simulation process $Litho$, network retraining process $Retrain$, maximum number of retraining rounds $R$.
    
    \For{$i$ = 1 to $R$}
    \State $adv\_train = Attack(F, D_{train, hotspot})$ \Comment{Generate adversarial non-hotspot layout for training hotspot.}
    \State $adv\_train\_hotspot = Litho(adv\_train)$ \Comment{Get verified hotspot through lithography simulation.}
    \State $D_{train} = D_{train} \cup adv\_train\_hotspot$
    \State $F = Retrain(F, D_{train})$ \Comment{Retrain network $F$ with robust training data $D_{train}$.}
    \EndFor
    \State \textbf{Return}: Robustified network $F$ 
    \end{algorithmic}
    \end{small}
    \end{algorithm}
    %\end{comment}

    % \todoblockgreen{Add a diagram to illustrate the principle of re-training i.e., doing a sort of pre-emptive/proactive attack on one's own model first...}

    \subsection{Evaluation}
        To demonstrate robustification, we perform iterative adversarial retraining on Network A, and perform white-box attacks to determine the attack success rate.
        % \todoblock{Details of retraining experiment}
        We start with Network~A in \autoref{sec:attack-methods}. We conduct the white-box attack using all hotspot layouts from the training set that are correctly classified by the CNN (\circled{1} of \autoref{fig:adv_retrain}).
        Of these 2070 hotspot layouts, the white-box attack successfully produces 1725 adversarial layouts (these were verified using lithography simulation, \circled{2} of \autoref{fig:adv_retrain}).
        These are labelled as hotspot, and added into the training dataset for the 1st round of retraining (\circled{3} of \autoref{fig:adv_retrain}).
        We call the 1st round retrained Network A'.
        For the second round of retraining, we take the hotspot layouts from the expanded training dataset that are correctly classified by Network A', and perform the white-box attack. 
        A' classifies 3910 hotspot layouts correctly, and from these, the white-box attack produces 2141 lithography simulation-verified hotspot layouts. These layouts are added into the training dataset and we perform a second round of retraining to produce the next generation of robustified network, Network A''. One can repeat this until an attack threshold success rate is met, or after a pre-determined number of rounds.

        To evaluate the efficacy of adversarial retraining, we perform white-box attacks using correctly classified hotspot layouts for Network A, A' and A'' from the validation dataset, after each round of training. When we use the white-box attacks to produce new training data and evaluate the retrained networks, we set the maximum number of SRAF insertions allowed ($T$) to 20, and the check parameter ($n$) to 180 (as in \autoref{sec:attack-results}).
        The results are shown in Table \ref{tab:retrain-result}.
        Throughout the retraining process, the networks' overall accuracy (the average of hotspot and non-hotspot classification accuracy) and hotspot detection accuracy are maintained.
        Even though Network A has a simpler architecture, after two rounds of retraining, its resilience surpasses that of Network B (see \autoref{tab:results}).

    \begin{table}[t]
    \caption{White-box attack results with iterative adversarial retraining on Network A.
    The maximum number of SRAF insertions allowed ($T$) is 20, and the check parameter ($n$) is 180. 
    }
    \label{tab:retrain-result}
    \begin{tabular}{@{}lrrr@{}}
    \toprule
    Network                  & Initial net (A) & 1st round retrain (A') & 2nd round retrain (A'') \\ \midrule
    Network overall accuracy         & 0.73        & 0.73              & 0.73              \\
    Hotspot detection accuracy         & 0.72        & 0.72              & 0.72              \\
    Attack success rate      & 99.7\%      & 73.6\%            & 37.2\%            \\
    Average attack time per layout    & 8.6 s       & 18.2 s            & 22.9 s            \\
    % Max num of pert. allowed & 20          & 20                & 20                \\
    % Check parameter          & 30          & 30                & 30                \\
    Average number of SRAFs added & 5.3         & 7.3               & 7.2               \\
    Average area of SRAFs added & 0.3\%       & 0.4\%             & 0.4\%             \\ \bottomrule
    \end{tabular}
    \end{table}

\section{Discussion}
    \label{sec:discussion}
    Over the course of this study, our findings raised several questions that warrant discussion and future study.
    % We discuss these peculiarities and some of the limitations of our study in this section.
    % ... and the wider implications to ML in CAD. We also discuss the limitations of this study, presenting opportunities for further exploration.
    
    % paragraphs, or subsections?

    \paragraph{What drives differences between white-box vs. black-box attack success rate?}
        Black-box attacks had a higher success rate compared to white-box attacks, at the cost of longer attack time. 
        We posit that this is largely due to the greedy nature of the black-box attack, where the target network is repeatedly queried.
        This query-based approach guarantees that the attacker can achieve the highest attack success rate (given a fixed horizontal and vertical sliding stride) in searching for the best shape/position combination.
        Conversely, while the white-box attack is a gradient-guided approach, it considers $n$ candidates and it is possible that the best valid solutions are missed depending on the size of the check parameter.
        
    \paragraph{What factors affect differences in robustness between shallower and deeper networks?}
        An interesting finding is that that there was a difference in attack success rate against Network A and B, where the more complex Network B displayed greater robustness even while the networks' baseline hotspot detection accuracies were the same.
        Network B's greater attack resilience is also supported by the lithography simulation-based verification, where less adversarial layouts were verified as still hotspot.
        Prior work has found that networks with greater capacity are more robust \cite{madry_towards_2017}; whether Network B has learned a "better" approximation of the underlying physics warrants more study.
            
    \paragraph{Do adversarial attacks generalize to other datasets?}
        Our study focuses on SRAF insertion for improving the printability of via layouts, which represents only one scenario in lithographic hotspot detection. Unfortunately, as noted previously, we were limited in our ability to comprehensively evaluate our attack on the ICCAD'12 contest benchmarks due to the unavailability of a DRC deck and lithography simulation settings for these benchmarks. Nevertheless, in the appendix, we show results for a limited evaluation of adversarial perturbation attacks on the ICCAD'12 dataset using a small set of inferred design rules.
        %and showed that our attack is equally successful. 
        Note that although we could not verify via lithographic simulation that the adversarially perturbed layouts remain hotspots, the perturbations are relatively small and are therefore unlikely to have actually fixed hotspots (as we have observed in the SRAF case study). Interestingly, the baseline CNN for the ICCAD'12 benchmarks has $>90\%$ accuracy, suggesting that higher accuracy by itself does not necessarily imply greater adversarial robustness.
        
       % To gauge broader applicability of our attack, we also examined another hotspot detection dataset (without lithography simulation-based verification) --- we share the details and results of that experiment in the appendix.
    \paragraph{What does the network learn?}
        Security aside, perhaps the more basic question raised by our work is this: what does a CNN learn?
        The high success rate of the attacks indicate that the CNN-based hotspot detectors do not truly and fully "learn" the physics relevant to the hotspot problem. One might argue that this is to be expected; after all, a CNN is only approximating the underlying physics. Nonetheless, the fact that in several cases only one or two additional SRAFs throws off the CNN is worrisome since, at least intuitively, these small modifications should not drastically fix hotspots.
        Indeed, in the broader deep learning community, there is on-going work about the interpretability of neural networks that seeks to better understand what concepts networks actually learn, and we would encourage this to be considered in the ML-CAD context also~\cite{montavon2018methods}.
        
    \paragraph{Are there wider security implications for ML-based CAD flows?}
        The success of both attacks in lithographic hotspot detection raises important questions about the wider implications to ML in CAD.
        CAD flows involve many complex steps using tools sourced from different vendors; this provides a wide attack surface \cite{basu_cad-base:_2019}. 
        With ML added to the mix, the risk compounds --- prior work has raised concerns about outsourcing deep learning \cite{gu_badnets:_2019}, and given the many CAD domains in which deep learning can contribute (we provide an overview in \autoref{sec:related}), security considerations are paramount.
        A key insight we provide in this study is the presence of semantically meaningful perturbations in the lithographic context. 
        We posit that similar meaningful perturbations exist in other CAD domains, and further work should be done to discover these.
    
\section{Related Work}
\label{sec:related}
    The CAD industry is facing challenges with increasing design complexity, especially with the growing time-to-market pressures.
    ML techniques have been explored to accelerate steps in the VLSI design flow \cite{kahng_machine_2018}. %; a major benefit of deep learning for CAD is speeding up the design flow.% in some way.
    
    In optical lithography, a variety of techniques have been proposed to analyze the printability of layouts and design enhancement.
    Pattern matching (such as \cite{yu2012accurate}) and ML (such as \cite{matsunawa2015new}) have been studied, offering a range of successes in accuracy and ability to generalize for previously unseen layouts. Recently, there has been an uptick in the study of deep learning approaches, where different facets have been investigated. For example, in \cite{yang_layout_2018}, Yang et al. train a CNN to detect hotspots from a layout image. They provide a detailed comparison on the effectiveness of different ML techniques at identifying hotspots, concluding that the CNN-based approaches offer superior accuracy. To reduce the computational overhead of processing large layout data, \cite{jiang_efficient_2019} proposes to binarize and down-sample the input data, yielding 8$\times$ speed-up over prior deep learning solutions.
    
    Other studies have exposed challenges in adopting CNNs, such as the abundance (or lack thereof) of labeled data for training.  Chen et al. detect hotspots using a CNN \cite{chen_semi-supervised_2019}, but propose a semi-supervised approach to handle the scarcity of labelled data. Using a two-stream architecture, labelled data is used to create a preliminary model that is used to provisionally label other samples together with a measure of confidence in the provisional label.  
    Provisionally labelled samples with high confidence are used to train the model in subsequent training cycles;
   % with a threshold for inclusion into the training set determined by the accuracy of networks trained from the provisionally labelled samples. 
   the general belief is that more data can result in ML models with more knowledge. 
   Synthetic variants of labelled data are proposed in \cite{reddy_enhanced_2018} to increase the size of the training dataset.
   The adversarial retraining procedure proposed in our work can similarly be viewed as a data augmentation strategy; however, unlike prior work, our data augmentation is targeted towards generating adversarially robust networks.
    
    Recently, state-of-the-art applications of CNNs have moved beyond design analysis towards design \textit{enhancement} to aid in modifying designs to reach a certain goal. Insertion of SRAFs has been framed as a type of image domain transformation, where Generative Adversarial Networks (GANs) are trained to take in layouts and "predict" where SRAFs should be inserted  \cite{alawieh_gan-sraf:_2019}. 
    Other mask optimizations (such as OPC) have been cast similarly \cite{yu_deep_2019, yang_gan-opc:_2018}.
    While they focus on accuracy and scalability, our work examines an orthogonal, yet crucial dimension of robustness.
    In physical design, trained ML models are a faster alternative to simulation, allowing designers to quickly evaluate the validity of a design. Lin et al. perform resist modelling and demonstrate transfer learning for different technologies \cite{lin_data_2018}. 
    Cao et al. \cite{cao_learning-based_2019} use parameters related to design, pin-mapping, and layout to predict achievable and actual inductance at pre- and post-layout stages. %. 
    
    Checking design rule violation (DRV) is another important aspect of the design flow where deep learning has been used. In \cite{tabrizi_machine_2018}, Tabrizi et al. do routability checks after netlist placement, but before global routing. Routing shorts are predicted using the trained model, allowing designers to avoid potential unroutable layouts. Similarly, Xie et al. \cite{xie_routenet:_2018} use a CNN to predict the number of DRVs, even in the presence of design macros and to identify DRV hotspots. DRV prediction ascertains the routability of layouts for earlier correction. 
    
    In early stages of design, deep learning has been used for logic optimization \cite{haaswijk_deep_2018}, design space exploration \cite{greathouse_machine_2018}, synthesis flow exploration \cite{yu_developing_2018}, and high-level area estimations \cite{zennaro_machine_2018}.
    Such techniques reduce designer workload by culling the design variants that need to be progressed in the design flow.  Yu et al. \cite{yu_developing_2018} propose the training and use of a CNN to gauge the effectiveness of different combinations of synthesis transformations (termed \textit{flows}) for a given register transfer level (RTL) design.
    To avoid exhaustively running all combinations, a model is trained from a \textit{subset} of possible flows.
    The trained model is used to predict the quality of several possible flows, outputting a collection of "angel-flows" that are likely to yield good results in the synthesis of the RTL design.
    In a related approach, Hasswijk et al. \cite{haaswijk_deep_2018} train a CNN to discover new transformation algorithms, but focus on graph optimization instead of quality of results from subsequent technology mapping.
    Orthogonal to these approaches, Greathouse et al. \cite{greathouse_machine_2018} use a neural network to predict how the performance of a software kernel will scale as a function of  the number of parallel compute units. %These performance estimates of new kernels at various points before committing the design in hardware.
    The adversarial robustness of these aforementioned approaches remains an open question.
    
    Adversarial attacks on CNNs are being actively studied \cite{biggio_wild_2018}, although, prior to our paper, this question of adversarial robustness has not yet been brought up in the electronic CAD domain. Our work contributes to this growing body of literature in this hitherto unanalyzed context.
    Our attacks modify layouts to cause misclassification on a well-trained network; these belong to the class of \textit{inference-time} or \textit{evasion} attacks and have been examined in detail (in a general context) in works like \cite{GoodfellowSS14, kurakin_adversarial_2016, szegedy_intriguing_2013, moosavi-dezfooli_universal_2017, sharif2016accessorize,Eykholt_2018_CVPR}.
    
    Adversarial attacks in the literature can be classified into two categories based on the contextual meaning/imperceptibility of the added perturbation. One class of adversarial perturbations are meaningless and akin to subtle noise that are crafted to fool the neural network in general cases. The other type of adversarial perturbations have contextual meaning, while still remaining subtle in the semantics of the real-world context. Examples of these include using a pair of glasses to mislead a face recognition system \cite{sharif2016accessorize} and a post-it note on a traffic sign to fool a traffic sign detector \cite{Eykholt_2018_CVPR}. Our attack on hotspot detection falls into the second category, as our added SRAFs are semantically meaningful (SRAFs are real-world artifacts) and difficult to perceive as malign.
    
    A variety of defenses against adversarial inputs have been proposed~\cite{stochastic_pruning, tramer_ensemble_2017, madry_towards_2017,papernot_distillation_2016}. Some focus on the detection of adversarial inputs by identifying feature disparity between valid and adversarial examples \cite{meng2017magnet,featuresqueezing,metzen2017detecting}. Some resort to transformation techniques to rectify adversarial inputs into "normal" ones \cite{meng2017magnet,defense-gan,counteering}. Others resort to retraining to counter adversarial attacks \cite{GoodfellowSS14, tramer_ensemble_2017}. Research into defenses that offer strong guarantees of robustness is ongoing \cite{madry_towards_2017}. %Aiming to maintain the quality and simplicity of inference of the original hotspot detection, 
    We use adversarial retraining to make CNN-based hotspot detectors robust without sacrificing detection accuracy and adding computation overhead in inference. 

    In contrast to \textit{inference-time} attacks, another class of attacks on deep learning is that of \textit{training-time} or \textit{backdoor} attacks \cite{gu_badnets:_2019, Trojannn}. Here the training dataset is in some way compromised (or poisoned).
    Part of the study into these attacks includes examining the risks when the integrity of the training data is compromised \cite{gu_badnets:_2019}, and the risks that come from re-using potentially compromised networks.
    Recent work that aims to improve the resilience to backdoors include \cite{liu_fine-pruning:_2018, neural_cleanse, chen_detecting_2019}.
    Understanding the implications of these attacks in ML-based CAD merits investigation.
    
\section{Conclusion}
\label{sec:conc}
    % \todoblock{%
        In this paper, we revealed a vulnerability of CNN-based hotspot detection in electronic CAD. We showed that CNN-based hotspot detectors are easily fooled by specially crafted SRAF insertions that can mislead the network to predict a hotspot layout as non-hotspot. We proposed and examined white-box and black-box attacks on well-trained hotspot detection CNNs, and the results showed that up to 99.7\% attack success rate was possible. The deeper, more complex CNN we attacked exhibited better natural robustness compared to the less complex CNN. To robustify the vulnerable hotspot detectors, we proposed adversarial retraining, revealing that after only two rounds, the white-box attack success rate could be decreased to 37.2\%.
        Our findings point to semantically meaningful adversarial perturbations as a viable concern for ML-based CAD. 
        This study leads us to urge caution and advocate for further study of the wider security implications of deep learning in this field. As an immediate recommendation for CNN-based hotspot detection, we suggest adversarial retraining as an add-on procedure after initial network training, as it introduces no extra computation overhead at inference and has no accuracy compromise, but adds robustness against adversarial attacks. 
        Ultimately, we find that adversarial perturbations are \textit{not necessarily} a showstopper for ML-based CAD, assuming, of course, that an appropriately proactive stance is adopted.
        Hence, our future work will look to other attack types in other CAD problems, including training-time attacks and robustification techniques.
    % }

% \clearpage
\bibliographystyle{ACM-Reference-Format}
\bibliography{sample-base}

\clearpage
\appendix
\section*{Appendix: Exploration using ICCAD '12 Dataset}

To further explore wider security implications in ML for CAD, we investigated adversarial perturbations in another hotspot detection task.
In this experiment, we attacked a different CNN-based hotspot detector, trained using a dataset from the ICCAD 2012 contest on pattern matching for physical verification \cite{iccad-dataset}.
The attack goal is similar to that in \autoref{sec:prelim} in that the attacker wants to modify given hotspot layouts such that they are misclassified as non-hotspot by the detector.
Note however, we could not perform lithography simulation-based verification as the physical models required were not available as part of the ICCAD competition dataset.

    % \subsection{Experiment Details}
    \paragraph{Dataset and Hotspot Detector Design}
        We use layout 4 from the ICCAD dataset containing 4547 training and 32067 test layout images.
        4452 of the training samples are non-hotspot and the remaining 95 are hotspot. 
        Of the test samples, 31890 are non-hotspot and 177 are hotspot.
        Each layout image has dimensions of 1200 pixel $\times$ 1200 pixel, and has binary valued pixel intensities to represent the pattern to be printed.
        Each pixel corresponds to $1~nm^2$ of the layout.
        Before training and inference we preprocess the layout images using the same DCT filters as described in \autoref{sec:cnn-design} to obtain the DCT coefficients as input to the hotspot detector.
        The resulting input dimension is (12, 12, 32).
        We use a similar CNN architecture and training procedure as for Network A (\autoref{sec:attack-methods}) but with this modified input dimension --- the network parameters are shown in \autoref{tab:network-architecture-appendix}.
        The trained network has 98.5\% non-hotspot classification accuracy and 92.7\% for hotspot classification accuracy on the test data.
        
        \begin{table}[H]
            \centering
            \caption{Network Architecture.}
            \label{tab:network-architecture-appendix}
        \begin{tabular}{@{}llll@{}}
            \toprule
            Layer & Kernel Size & Stride & Output Size          \\ \midrule
             input          &   -   &   -   &   (12, 12, 32)    \\
             conv1\_1       &   3   &   1   &   (12, 12, 16)    \\
             conv1\_2       &   3   &   1   &   (12, 12, 16)    \\
             maxpooling1    &   2   &   2   &   (6, 6, 16)      \\
             conv2\_1       &   3   &   1   &   (6, 6, 32)      \\
             conv2\_2       &   3   &   1   &   (6, 6, 32)      \\
             maxpooling2    &   2   &   2   &   (3, 3, 32)      \\
             fc1            &   -   &   -   &   250             \\ 
             fc2            &   -   &   -   &   2               \\ \bottomrule
        \end{tabular}
        \end{table}
    
    % Layout image dimension (1200, 1200), 32 DCT filters, size (100, 100), stride 100, results in (12, 12, 32) as inputs to the hotspot detection network.

\begin{figure}[b]
            \centering
            \includegraphics[width=\textwidth]{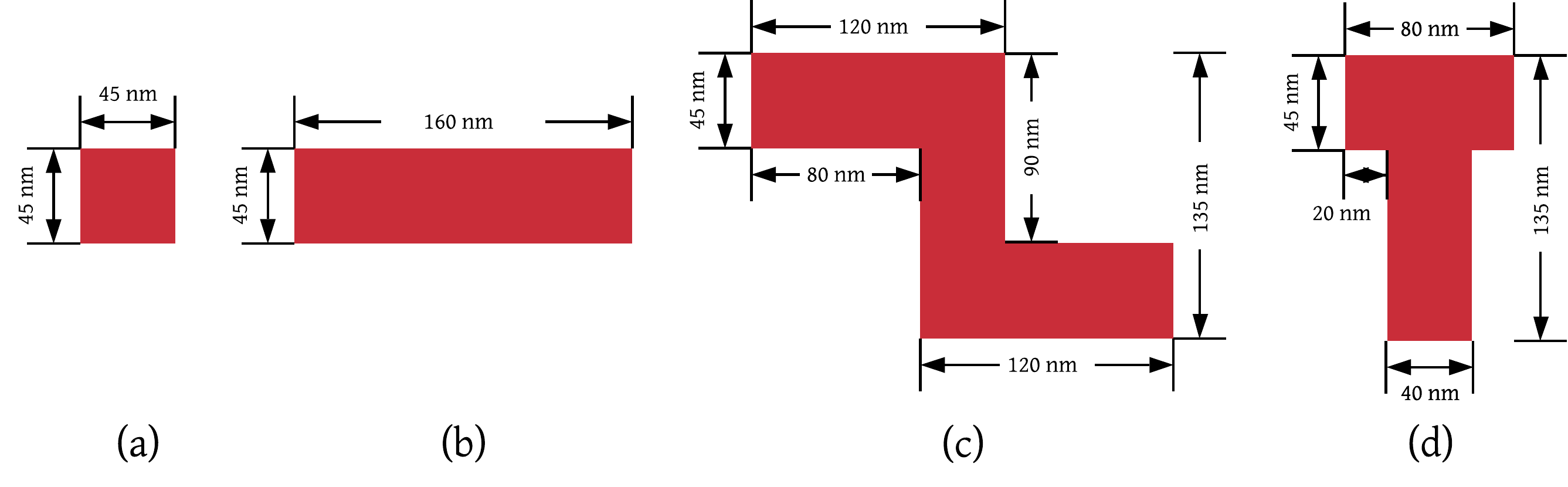}
            \caption{Restricted set of patterns for black-box attack (shape dimensions shown). The (a) square, (b) rectangle, (c) zig-zag, and (d) Tee shapes are drawn from the existing layout dataset.}
            \label{fig:appendix_shape}
        \end{figure}

    \paragraph{Black-box attack}
        We conducted a black-box attack on the test hotspot layouts to examine the efficacy of our proposed attack scheme.
        However, instead of inserting SRAFs, we add isolated printing patterns to the layouts.
        The attack constraints in this experiment are: (1) shape constraint: adversarial insertions can only be chosen from a restricted set of four basic shapes that already exist in the layout dataset, as illustrated in \autoref{fig:appendix_shape}; (2) spacing constraint: inserted patterns should be at least 45~nm away from any surrounding patterns; (3) alignment constraint: inserted patterns need to be aligned with existing shapes; (4) insertion region: inserted patterns must not overlap with a 100~nm wide border at the edges of the layout image.
        The black-box algorithm is reused from \autoref{sec:attack-methods}.

\begin{figure}[b]
        \begin{center}$
        \begin{array}{cccc}
        
        \includegraphics[width=0.22\textwidth]{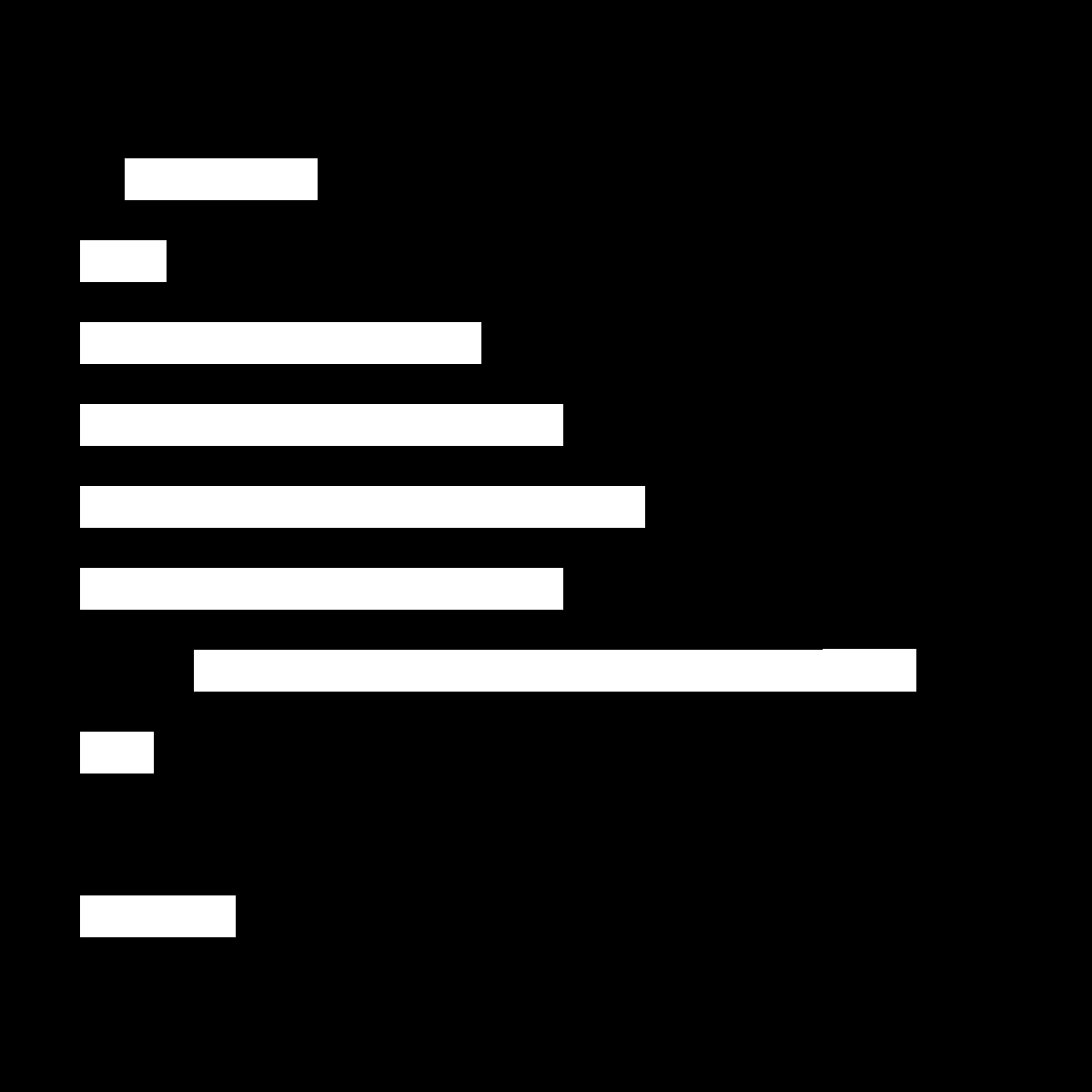}& 
        \includegraphics[width=0.22\textwidth]{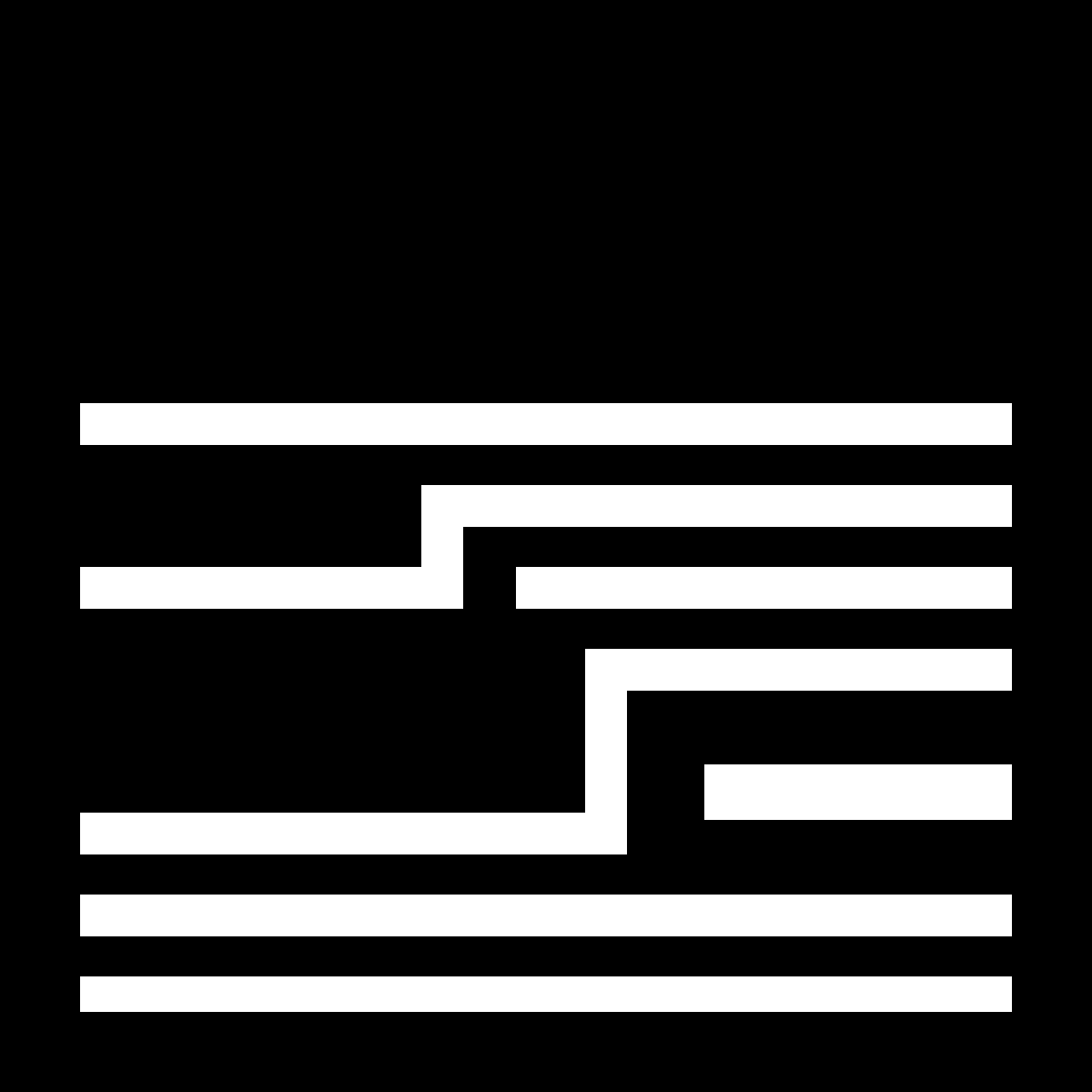}&
        \includegraphics[width=0.22\textwidth]{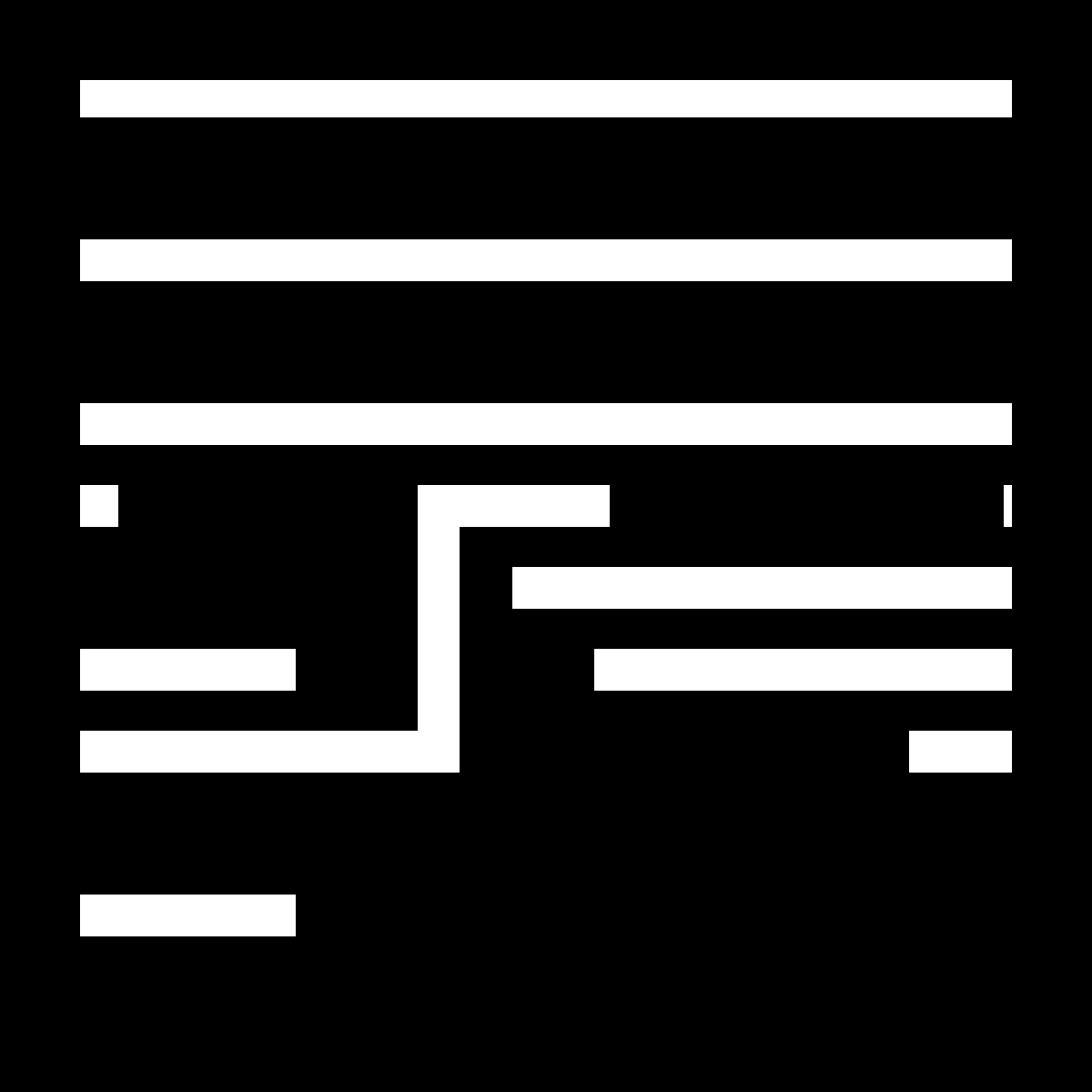}&
        \includegraphics[width=0.22\textwidth]{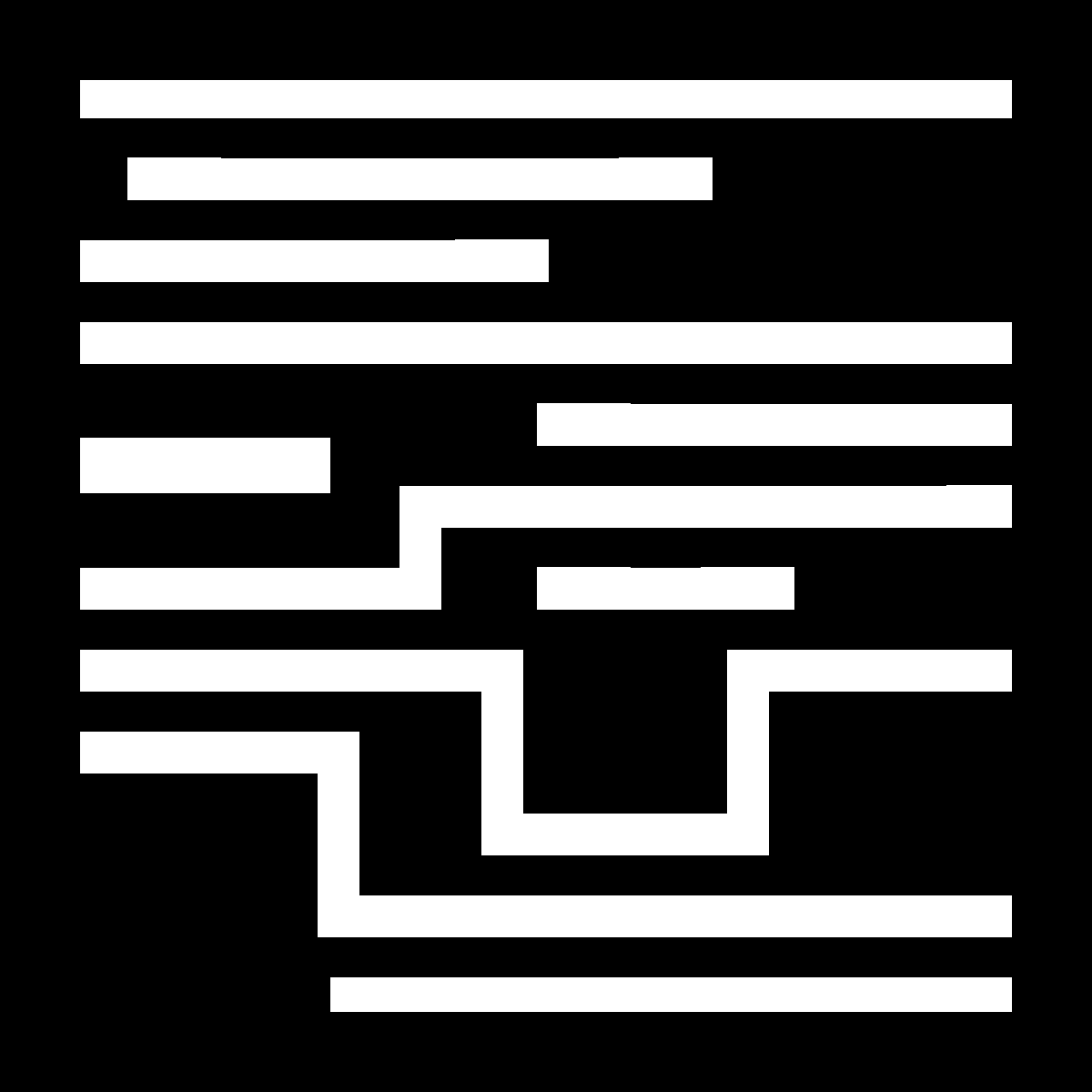}\\
        
        \includegraphics[width=0.22\textwidth]{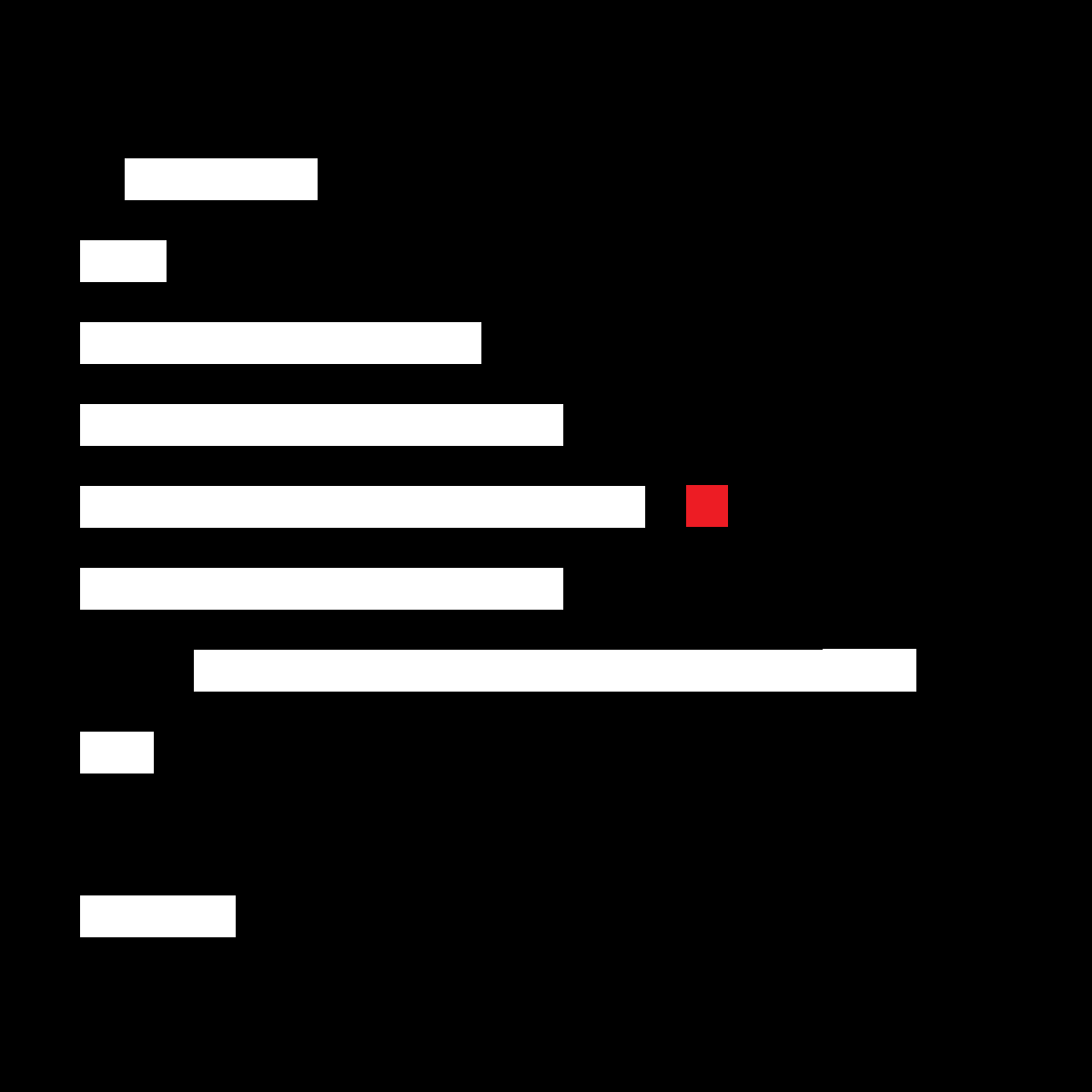}&
        \includegraphics[width=0.22\textwidth]{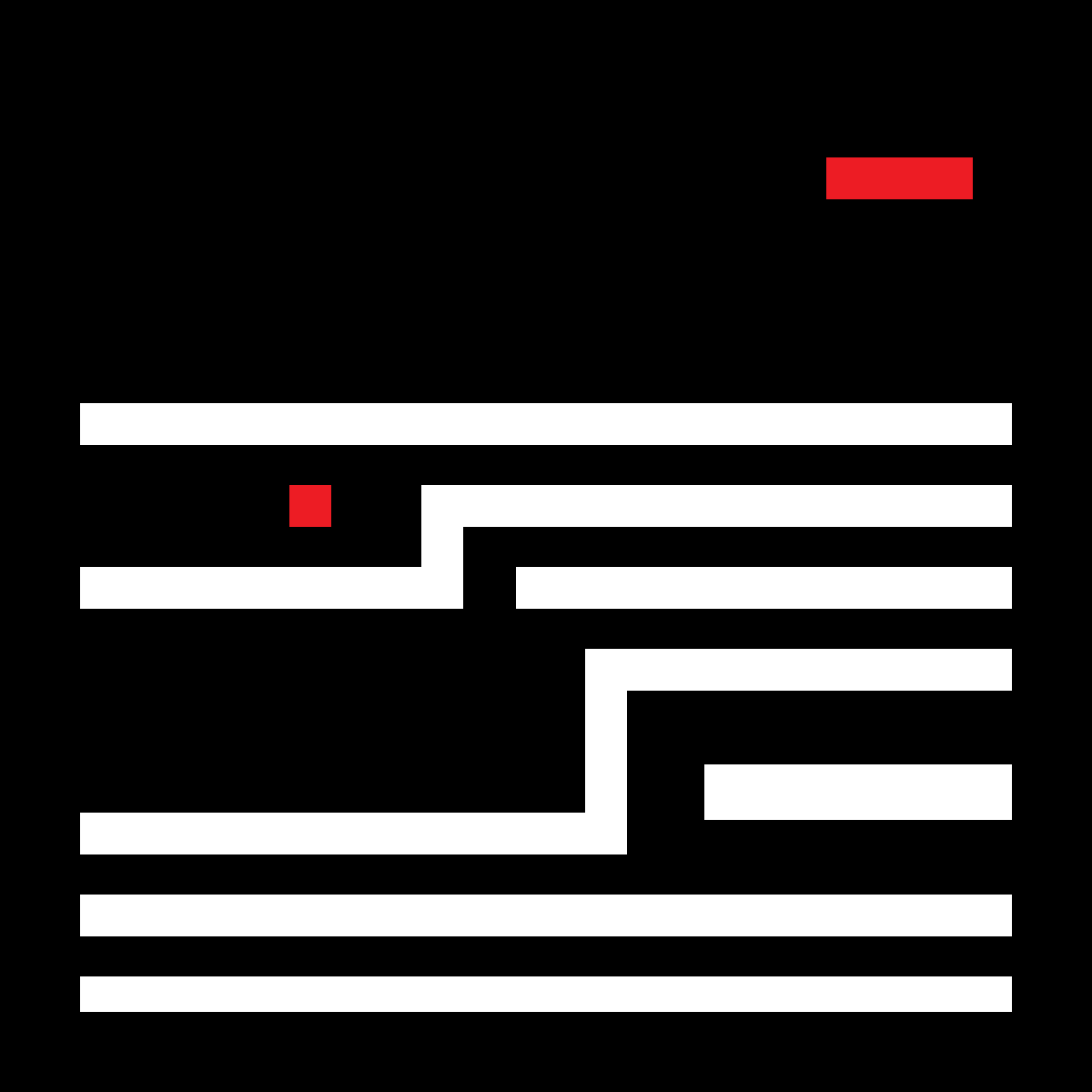}&
        \includegraphics[width=0.22\textwidth]{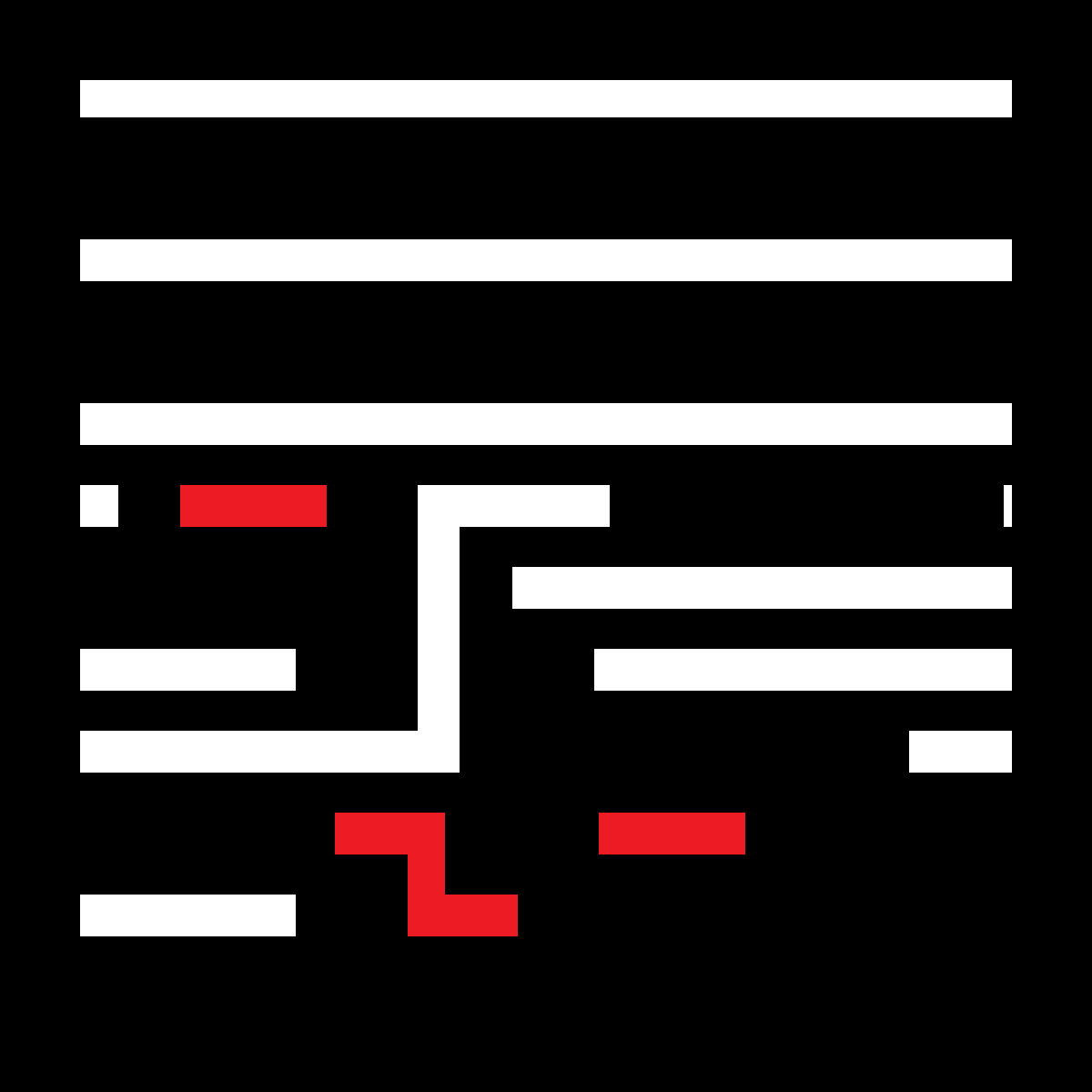}&
        \includegraphics[width=0.22\textwidth]{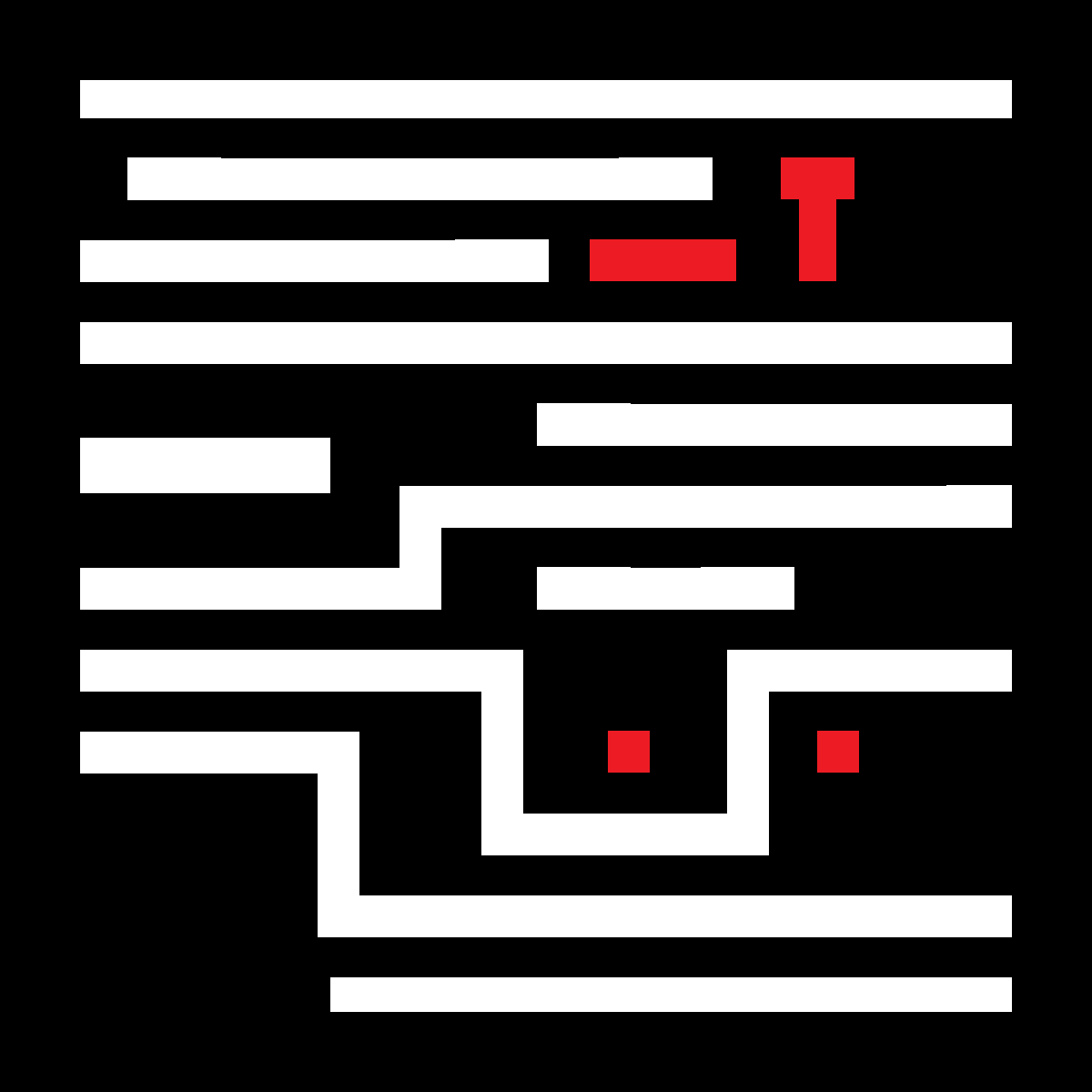}\\
        
        \end{array}$
        \end{center}
        \caption{Top row: original hotspot layouts. Bottom row: corresponding adversarial non-hotspot layouts.}
        \label{Fig:advhotspot-appendix}
        \end{figure}

    \paragraph{Attack Results}
        We performed the black-box attack using 164 hotspot layouts from the test set that the detector correctly classified as hotspot.
        Of these 164 layouts, 127 were successfully perturbed to fool the network.
        The results are summarized in \autoref{tab:appendix-results}.
        We illustrate some of the perturbed layouts in \autoref{Fig:advhotspot-appendix} with 1-4 adversarial inserted patterns.
    
        \begin{table}[t]
        \centering
        \caption{Summary of black-box attack result}
        \label{tab:appendix-results}
        \begin{tabular}{lrrrr}
        \hline
        Attack success rate         & 77.4\%    \\
        Average attack time per layout & 63.5 s    \\
        Average number of patterns added    & 4.5       \\
        Average area of patterns added    & 2.1\%     \\ \hline
        \end{tabular}
        \end{table}

    \paragraph{Remarks}
        Based on these additional experiments, it appears that this dataset is also susceptible to adversarial perturbation attacks, despite the fact that the CNN-based hotspot detector baseline accuracy on this dataset is high.
        Access to the lithography simulation settings will further allow us to verify that the modified layouts remain hotspots.

% 
% If your work has an appendix, this is the place to put it.

\end{document}